\documentclass{article}

\PassOptionsToPackage{numbers,compress}{natbib}

 \usepackage[main, final]{arxiv_neurips}

\usepackage[utf8]{inputenc} 
\usepackage[T1]{fontenc}    
\usepackage{hyperref}       
\usepackage{url}            
\usepackage{booktabs}       
\usepackage{amsfonts}       
\usepackage{nicefrac}       
\usepackage{microtype}      
\usepackage{xcolor}         
\usepackage{microtype}
\usepackage{graphicx}
\usepackage{subcaption}
\usepackage{booktabs} 
\usepackage{xcolor} 
\newcommand{\pos}[1]{\textcolor{green!55!black}{#1}}
\newcommand{\negdiff}[1]{\textcolor{red!70!black}{#1}}
\usepackage{algorithm} 
\usepackage{algpseudocode}
\usepackage{seqsplit}
\usepackage{fvextra}
\usepackage[dvipsnames]{xcolor}
\usepackage{float}
\usepackage{caption}
\usepackage{balance}

\usepackage{hyperref}

\usepackage{booktabs}
\usepackage{multirow}

\usepackage{amsmath}
\usepackage{amssymb}
\usepackage{mathtools}
\usepackage{amsthm}

\usepackage[capitalize,noabbrev]{cleveref}

\theoremstyle{plain}

\theoremstyle{definition}

\theoremstyle{remark}

\usepackage[textsize=tiny]{todonotes}
\usepackage[most]{tcolorbox} 
\usepackage{booktabs,multirow,makecell}
\usepackage{wrapfig}
\usepackage{graphicx}
\tcbset{fonttitle=\bfseries}

\makeatletter
\@ifundefined{theHALG@line}
  {\newcommand{\theHALG@line}{\thealgorithm.\arabic{ALG@line}}}
  {\renewcommand{\theHALG@line}{\thealgorithm.\arabic{ALG@line}}}
\makeatother

\title{ViSRA: A Video-based Spatial Reasoning Agent for Multi-modal Large
Language Models}

\author{%
  \textbf{Tingshu Mou}$^{1,*,\dagger}$ \quad
  \textbf{Jiabo He}$^{2,*}$ \quad
  \textbf{Renying Wang}$^{2}$ \quad
  \textbf{Ce Liu}$^{2}$ \\
  \textbf{Hao Yang}$^{2}$ \quad
  \textbf{Tiehua Zhang}$^{3}$ \quad
  \textbf{Jingjing Chen}$^{1}$ \quad
  \textbf{Xingjun Ma}$^{1,\ddagger}$ \\
  \\
  $^{1}$Fudan University \quad
  $^{2}$Bosch Center for Artificial Intelligence (BCAI) \quad
  $^{3}$Tongji University
}
\begin{document}

\maketitle
\begingroup
\renewcommand{\thefootnote}{}
\footnotetext{$^{*}$Equal contribution.}
\footnotetext{$^{\dagger}$The work was completed during Tingshu's internship at BCAI.}
\footnotetext{$^{\ddagger}$Corresponding author (xingjunma@fudan.edu.cn).}

\endgroup

\begin{abstract}
  Recent advances in Multi-modal Large Language Models (MLLMs) target 3D spatial intelligence, yet the progress has been largely driven by post-training on curated benchmarks, leaving the inference-time approach relatively underexplored. In this paper, we take a training-free perspective and introduce \textbf{ViSRA}, a human-aligned \textbf{Vi}deo-based \textbf{S}patial \textbf{R}easoning \textbf{A}gent, as a framework to probe the spatial reasoning mechanism of MLLMs. ViSRA elicits spatial reasoning in a modular and extensible manner by leveraging explicit spatial information from expert models, enabling a plug-and-play flexible paradigm. ViSRA offers two key advantages: (1) human-aligned and transferable 3D understanding rather than task-specific overfitting; and (2) no post-training computational cost along with heavy manual curation of spatial reasoning datasets. 
Experimental results demonstrate consistent improvement across a set of MLLMs on both existing benchmarks and unseen 3D spatial reasoning tasks, with ViSRA outperforming baselines by up to a $15.6\%$ and $28.9\%$ absolute margin respectively.
\end{abstract}

\section{Introduction}
Multi-modal Large Language Models (MLLMs) have recently shown impressive progress in following instructions grounded in visual inputs~\cite{brazil2023omni3d, zha2025enable, daxberger2025mm, li2025perception, liu2025spatial}, which naturally raises a broader question: \textit{can these models develop genuine 3D-centric spatial intelligence?} Spatial reasoning in 3D scenes—e.g., understanding relative directions and distances among objects, requires more than visual perception and semantic awareness~\cite{yang2025thinking}. It demands building and manipulating an intermediate representation of space that is consistent across viewpoints and time. This is particularly critical for real-world 3D scenes (e.g., videos), where the model must maintain object permanence and track evolving relations under camera motion with human-aligned perceptual routines. 
Despite impressive performance on 2D tasks, current MLLMs struggle to essentially understand the 3D space and perform poorly on existing benchmarks.

\begin{figure*}[t]
  \begin{center}
    \includegraphics[width=0.9\textwidth]{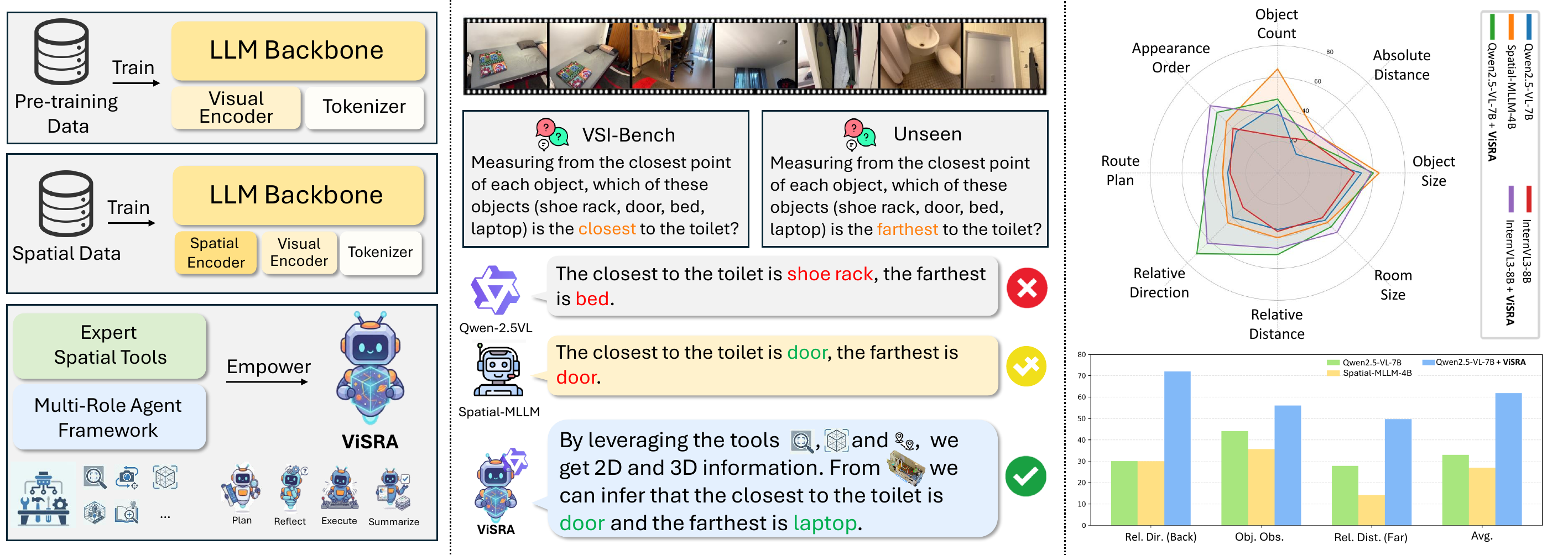}
    \caption{Comparison of three paradigms for 3D spatial reasoning. We evaluate ViSRA against the pre-trained base MLLM (e.g., Qwen-2.5VL) and the post-training method (e.g., Spatial-MLLM). Left: three paradigms for 3D spatial reasoning. Middle: qualitative examples showing that the base model can fail on both established and unseen questions, the post-trained model can succeed on established but fail on unseen questions, while ViSRA succeeds on both. Right: quantitative comparison across established and unseen tasks, where ViSRA achieves the best overall performance.}
    \label{fig:head}
  \end{center}
\end{figure*}

A common strategy to address the above problem is to post-train MLLMs on spatial datasets via supervised fine-tuning, architectural modifications, instruction tuning, or preference-based optimization~\cite{batra2025spatialthinker, yang2025visual, cai2025scaling, li2025spatialladder, wu2025spatial, fan2025vlm}. While this line of works can boost accuracy on curated spatial suites, it also introduces several concerns. 
Firstly, the post-training pipeline involves manual collection and curation of large-scale video reasoning datasets and comes with high computational cost. 
More fundamentally, reliance on existing spatial datasets can encourage benchmark-specific overfitting, producing apparent improvements that may not transfer to out-of-distribution (OOD) problems~\cite{yang2025cambrian, lin2025mmsi}, as shown in Figure~\ref{fig:head}(middle).
In real-world scenarios which are often far more complex than curated benchmark tasks, such methods may neither reliably solve the problem nor provide cues that align with human reasoning.

To address these issues, we propose \textbf{ViSRA}, an inference-time, human-aligned \textbf{Vi}deo-based \textbf{S}patial \textbf{R}easoning \textbf{A}gent that enhances MLLMs’ spatial reasoning via the dynamic, modular integration of state-of-the-art visual perception tools and a multi-role framework. 
Specifically, we design seven types of tools that leverage domain-expert models to extract 2D and 3D object information, as well as scene geometry. ViSRA then adopts a four-role agent framework to exploit spatial information through planning, a reflection--execution loop, and final summarization.
This design yields two main benefits: (i) it directly inherits continual advances in perception models without additional post-training; and (ii) it is potentially a generalizable and human-aligned agent framework, where intermediate tracked objects, reconstructed geometry, and derived relations serve as explicit spatial cues for answering questions.
We evaluate on VSI-Bench~\cite{yang2025thinking} across multiple base MLLMs and observe consistent gains by up to $15.6\%$ in spatial intelligence. We further introduce VSI-Bench-Extra to assess generalization to unseen questions, where ViSRA consistently surpasses benchmark-specific post-trained models and outperforms base MLLMs by up to $28.9\%$.

To conclude, our contributions are as follows:
\begin{itemize}
    \item We empirically show that current MLLMs often lack reliable 3D understanding and struggle with robust and generalizable 3D spatial reasoning, particularly under distribution shifts.
    \item We introduce ViSRA, an inference-time video-based spatial reasoning agent that enhances the spatial intelligence of existing MLLMs. ViSRA dynamically leverages continually improving perception models to construct explicit spatio-temporal evidence in a human-aligned manner, achieving strong performance without post-training on  spatial benchmarks.
    \item We further observe that ViSRA performs superiorly on unseen spatial tasks, indicating promising generalization potential to OOD challenges and real-world corner cases.
\end{itemize}

\section{Related Work}
\label{sec: related}

\begin{wrapfigure}{R}{0.38\columnwidth}
  \centering
  
  \includegraphics[width=0.38\columnwidth]{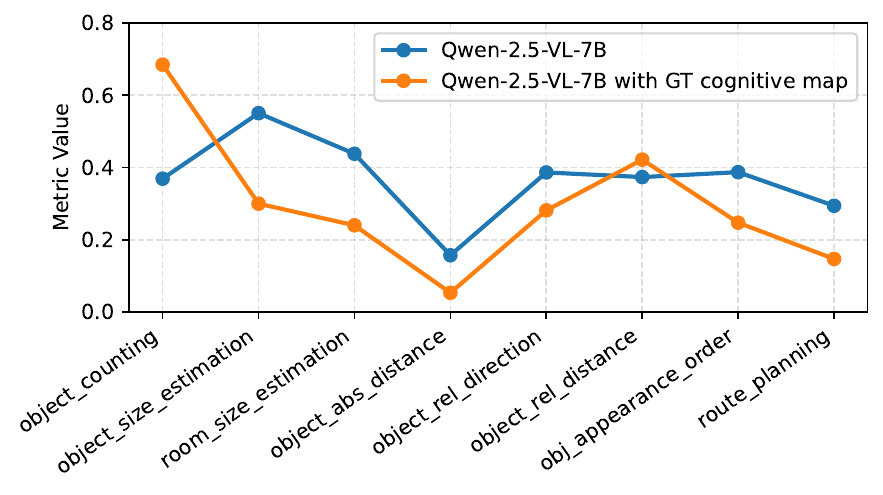}
  \captionof{figure}{Performance comparison across problem types on VSI-Bench (a subset of $779$ questions). Qwen2.5-VL-7B yields a drop on six question types given ground-truth(GT) cognitive maps.}
  \label{fig:lineplot_vsi_tasktype}
  
  \vspace{0.4em} 
  
  \includegraphics[width=0.38\columnwidth]{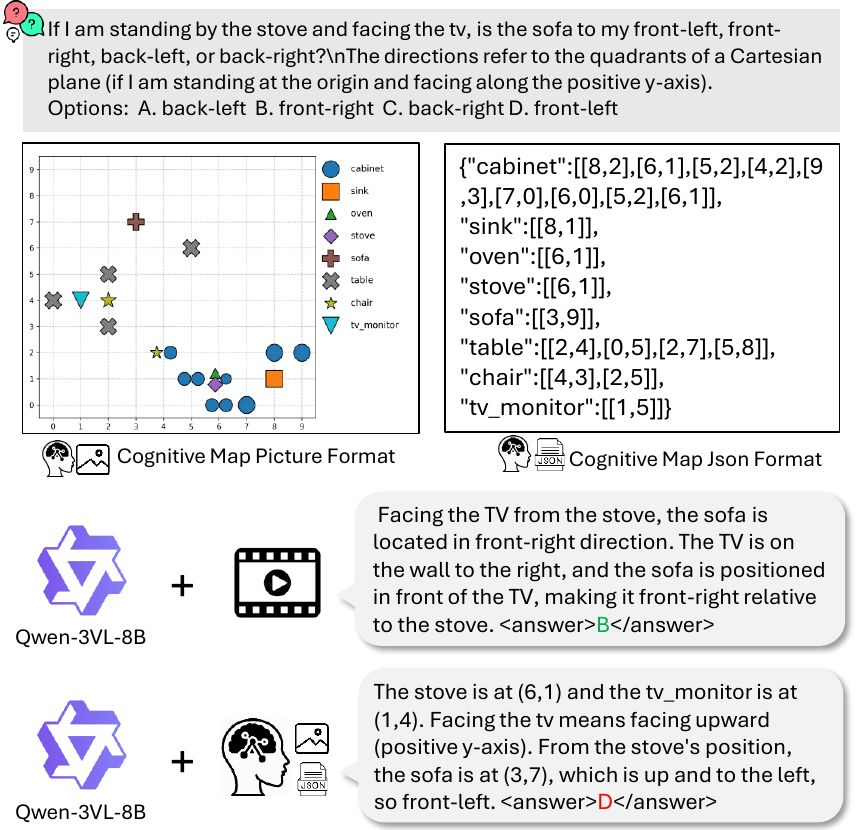}
  \captionof{figure}{A comparison example. Qwen3-VL-8B succeeds to answer a spatial question with the source video but outputs the wrong answer with the summarized cognitive map.}
  \label{fig:cogmap}
  \vspace{-1.2em}
\end{wrapfigure}
\subsection{MLLMs for Video Understanding}
MLLMs have made substantial progress in video understanding, exhibiting strong capability in modeling high-level semantics and temporal dynamics from video inputs.
Early video MLLMs~\cite{zhang2023video, fei2024video, maaz2024video, lin2024video} typically extended image-based vision and language models to videos by combining a pre-trained visual encoder with a language model, and by scaling video and text alignment as well as instruction tuning on large datasets.
Subsequent works further improved training and inference strategies, leveraging emerging techniques such as structured reasoning and reinforcement learning. For instance, Video-R1~\cite{feng2025video} adapted reinforcement learning to video understanding and proposed T-GRPO to better exploit temporal information and enhance reasoning over long videos.
MotionEpic~\cite{fei2024video} incorporated a spatio-temporal Scene Graph (STSG) as structured input and output and introduced a Video-of-Thought inference framework, enabling fine-grained video understanding and grounding.
VideoAgent~\cite{fan2024videoagent} proposed a multi-modal agentic framework that improved explainability and generalization by automatically invoking pre-defined tools.
Despite these advances, existing video MLLMs remained primarily optimized for semantic video understanding and often underperformed on video-based spatial reasoning tasks.

\subsection{Spatial Reasoning in MLLMs}
We have witnessed growing interest in spatial reasoning for MLLMs in recent years, accompanied by the emergence of benchmarks that systematically evaluated video-based spatial intelligence~\cite{yang2025thinking, li2025sti, wu2025st, lin2025mmsi}.
To our best knowledge, VSI-Bench was first proposed as a video-based visual-spatial intelligence benchmark to probe MLLMs' perceptual, linguistic, and temporal capabilities on spatial reasoning tasks~\cite{yang2025thinking}.
Following benchmarks, such as STI-Bench~\cite{li2025sti} and MMSI-Video-Bench~\cite{lin2025mmsi}, were also designed to evaluate MLLMs’ spatio-temporal understanding through challenging tasks, revealing MLLMs' limitations in real-world spatio-temporal understanding, ranging from spatial construction and motion understanding to planning, estimation, prediction, and cross-video reasoning.

To endow MLLMs with stronger spatial intelligence, a bunch of works adopted post-training with explicit spatial encoders~\cite{feng2025towards, wu2025spatial, chen2025reasoning, fan2025vlm, zheng2025learning} and large-scale spatially grounded data~\cite{yang2025visual, cai2025scaling, li2025spatialladder}.
Spatial-MLLM~\cite{wu2025spatial} integrated VGGT~\cite{wang2025vggt} into a dual-encoder design, exploiting the spatial priors provided by geometry foundation models.
GS-Reasoner~\cite{chen2025reasoning} targeted 3D visual grounding and spatial understanding via a dedicated fusion mechanism for geometric and semantic features, together with a Grounded Chain-of-Thought (GCoT) dataset that provided spatial reasoning traces.
On the data side, VST-P and VST-R~\cite{yang2025visual} scaled spatial supervision with a million-level dataset for enhancing spatial perception and a 135K-sample dataset for spatial reasoning instruction.
Cai \emph{et al.}~\cite{cai2025scaling} further scaled spatial training data to eight million diverse samples and reported strong performance across a broad range of spatial intelligence benchmarks.

In addition to post-training, training-free methods~\cite{liu2025coarse, qi2025gpt4scene} enhanced spatial reasoning by injecting signals from spatial expert models.
GPT4Scene~\cite{qi2025gpt4scene} reconstructed scenes and rendered Bird's Eye View (BEV) images as inputs to facilitate spatial reasoning.
Coarse Correspondences~\cite{liu2025coarse} leveraged tracking models (e.g., Tracking Anything~\cite{yang2023track}) to associate objects across frames and fed object-marked frames to MLLMs, improving spatio-temporal reasoning without task-specific fine-tuning.
While there have been a few training-free works, a human-aligned agent that improves MLLMs' spatial reasoning in a generalizable and explicit way is still underexplored.

\begin{figure*}[t]
  \begin{center}
    \includegraphics[width=0.9\textwidth]{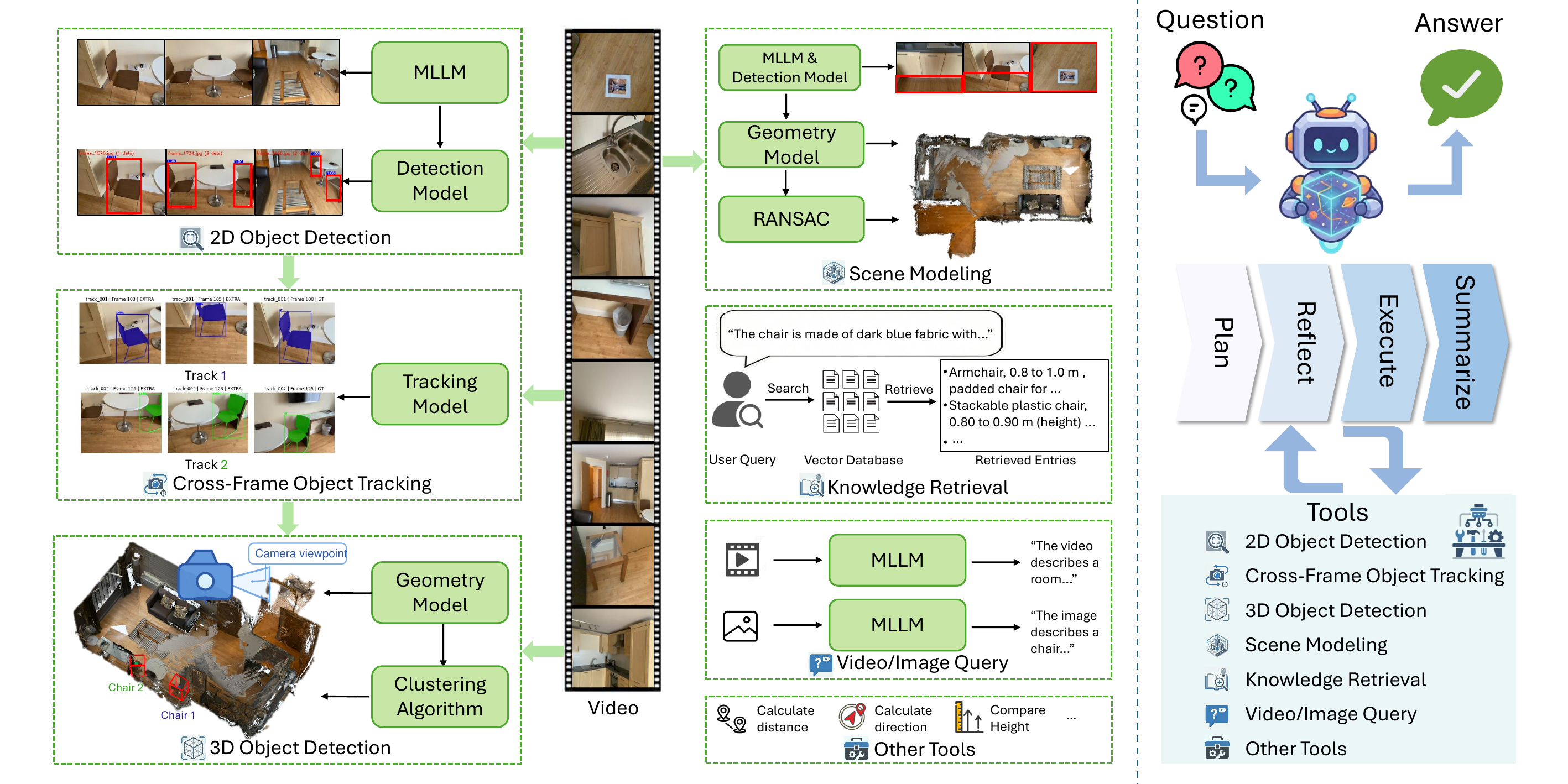}
    \caption{Overview of ViSRA. The left panel summarizes spatial tools that produce accurate intermediate predictions, while the right panel illustrates the multi-role agent framework that orchestrates these tools to solve spatial queries.}
    \label{fig:overview}
  \end{center}
\end{figure*}

\section{Proposed Approach}
In this section, we first present observations and analyses of current MLLMs on spatial reasoning in Section~\ref{sec: ob}, and then introduce our approach from Section~\ref{sec: overview} to Section~\ref{sec: agent}.

\subsection{Limitations of Current Approaches}
\label{sec: ob}


At the beginning of our exploration, we wondered what could enhance the spatial intelligence of MLLMs.~\citet{yang2025thinking} suggested that cognitive maps represented internal layouts of environments, serving as a potentially interpretable scaffold for improving spatial reasoning. We thus conducted an experiment on a subset of VSI-Bench~\cite{yang2025thinking} by providing each scene’s ground-truth (GT) cognitive map (generation details in Appendix~\ref{app:cognitive}) as an additional input. However, results in Figure~\ref{fig:lineplot_vsi_tasktype} show that cognitive maps were largely ineffective in practice across most task types. 
More surprisingly, performance even dropped on tasks that humans would typically regard as well supported by a map, such as inferring the relative direction among multiple objects. A qualitative example in Figure~\ref{fig:cogmap} highlights this limitation: the model answered correctly only with the original video, but failed once the cognitive map was provided. This indicates that existing MLLMs can miss basic spatial relations that are straightforward for humans, thus failing to own human-aligned capability.

As mentioned in Section~\ref{sec: related}, post-trained models (e.g., Spatial-MLLM~\cite{wu2025spatial}) achieved strong performance on VSI-Bench, but they were unable to produce human-aligned spatial reasoning cues. This raised the concern that such improvements might reflect benchmark-specific adaptation rather than transferable spatial understanding. To probe generalization, we constructed VSI-Bench-Extra by extending VSI-Bench with the same video sources and introducing three additional question types to evaluate MLLMs on out-of-distribution (OOD) spatial questions.
As shown in Figure~\ref{fig:head} (bottom right), post-trained models did not outperform the baselines, revealing that current post-training methods did not deliver generalizable 3D spatial reasoning.

Overall, it suggests two takeaways: (i) existing MLLMs still lack human-aligned capability of spatial reasoning; and (ii) post-training methods primarily boost in-benchmark performance, but yield negligible OOD generalization.

\subsection{Overview of Our Agentic Approach}
\label{sec: overview}
Motivated by the above explorations and observations, we propose \textbf{ViSRA}, an inference-time \textbf{Vi}deo-based \textbf{S}patial \textbf{R}easoning \textbf{A}gent framework in a human-aligned manner. As shown in Figure~\ref{fig:overview}, ViSRA consists of (i) a suite of spatial tools and (ii) a multi-role agent architecture. Spatial tools leverage expert models from relevant domains to extract visual and geometric cues. Given video and question prompts, ViSRA produces human-aligned spatial reasoning by orchestrating tool calls and aggregating their results in four roles. We describe relevant spatial tools in Section~\ref{sec: spatial tools} and the agent design in Section~\ref{sec: agent}.

\subsection{Spatial Tools}
\label{sec: spatial tools}
As illustrated in Figure~\ref{fig:overview}, we implement a suite of spatial tools that can be invoked by ViSRA. These tools expose object-centric temporal dynamics and scene-centric geometric structures, enabling ViSRA to retrieve where an object appears over time and spatial relationships among objects in a consistent 3D space. Concretely, the tool suite includes 2D object detection, cross-frame object tracking, 3D object detection, scene modeling, knowledge retrieval, video/image query, and other utility tools for spatial reasoning.

\begin{figure*}[t]
  \begin{center}
    \includegraphics[width=0.9\textwidth]{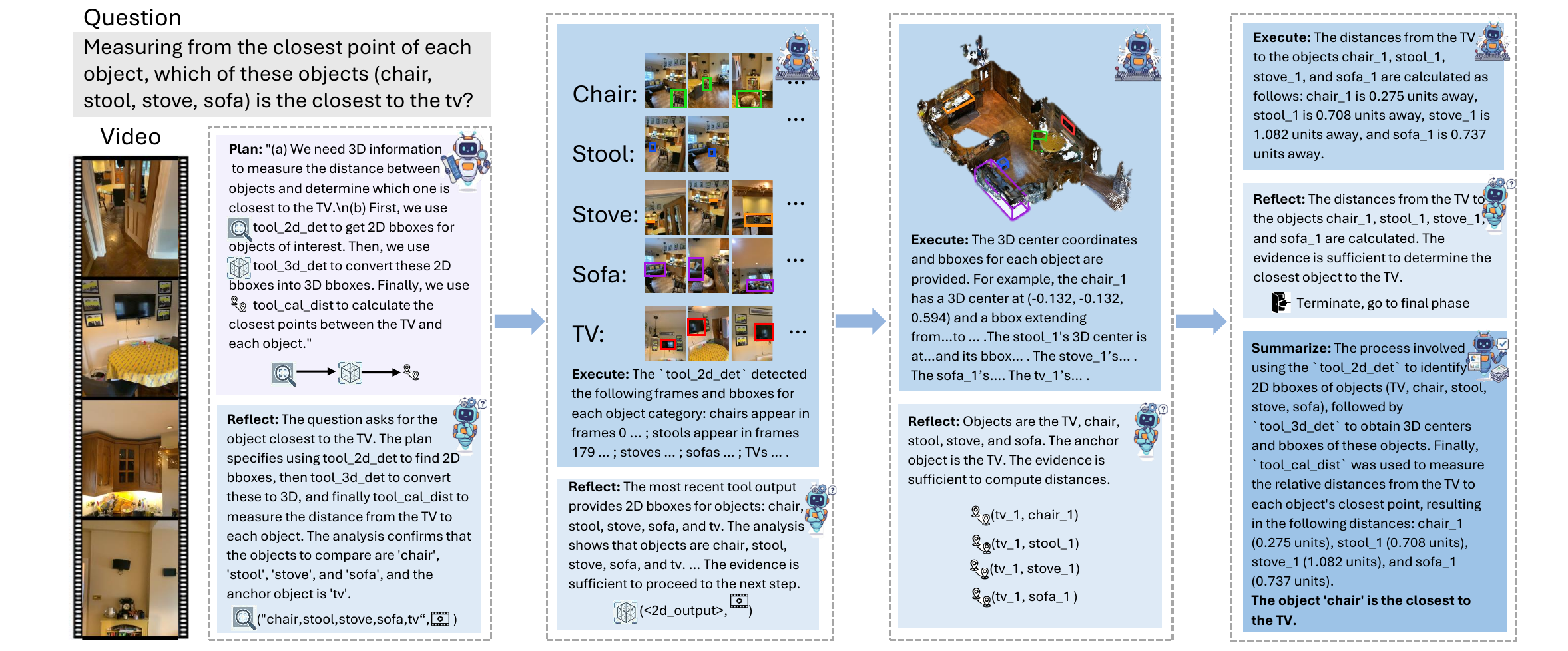}
    \caption{An inference example. ViSRA answers a relative-distance question by using four roles and invoking multiple spatial tools.}
    \label{fig:example}
  \end{center}
\end{figure*}

\paragraph{2D object detection.}
When answering video-based spatial questions about objects, a natural first step is to identify key frames in which the target objects appear and localize their image regions. This tool performs per-frame object detection on key frames, treating each detection as a view of the target object. Given a set of object queries from a question, the MLLM first performs frame filtering by prompting each sampled frame with ``Does this frame contain the following objects: \ldots?'' to select candidate key frames. An image detection model (e.g., Rex-omni~\cite{jiang2025detect}) is then applied to the selected frames to generate bounding boxes (bboxes) for target objects. The tool returns the indices of the selected frames along with corresponding bboxes.

\paragraph{Cross-frame object tracking.}
Humans track an object across time using appearance cues such as shape and color. Following this intuition, we apply a tracking model (e.g., Segment Anything Model (SAM)~\cite{carion2025sam}) to propagate 2D bboxes from the detection tool across frames. This produces tracklets that associate multiple views of the same object instance throughout the video.

\paragraph{3D object detection.}
Beyond appearance cues, humans also rely on geometric consistency to recognize object instances and distinguish their locations. To provide such spatial cues, we use a foundational geometric model (e.g., VGGT~\cite{wang2025vggt}) for object-level 3D detection, mapping 2D pixels inside each detected bbox to 3D points, and computing the 3D center of each view as the centroid of these 3D points.
We then cluster different views of the same object instances based on Euclidean distances between their 3D centers. Details of our proposed clustering algorithm are provided in Appendix~\ref{Clustering Algorithm}. Camera information for each frame can also be calculated for better 3D understanding.

\paragraph{Scene modeling.}
Humans form a coherent scene modeling mentally while watching a video. Although the foundational geometric model reconstructs geometry as point clouds, it does not explicitly provide a stable ground-plane reference. We therefore estimate the ground plane via Random Sample Consensus (RANSAC)~\cite{fischler1981random}. Concretely, we detect ``floor'' regions using the 2D detection tool and lift the corresponding pixels to 3D using the 3D detection tool. We then fit a plane to these 3D points via RANSAC and average the estimated plane parameters across frames to obtain a robust ground-plane estimate. The plane normal defines the vertical axis, and we set the positive direction (i.e., ``up'') as the half-space containing the majority of reconstructed points. This yields a real-world-aligned scene coordinate frame, enabling view transformations such as the bird's-eye-view (BEV) rendering.

\paragraph{Knowledge retrieval.}
Some estimation questions require spatial priors beyond visual evidence. To address this, we introduce a knowledge retrieval procedure that provides object- and scene-level size statistics. We compile a knowledge file with more than $500$ entries of common objects and rooms from real-world sources, each with size statistics and a brief description. Given the question, the MLLM retrieves the five most relevant entries and uses them as additional priors to answer otherwise ambiguous estimation queries. This human-aligned design also enables us to understand the role of spatial priors in estimation tasks.

\paragraph{Video/Image query.}
The agent can also directly query the video (or selected key frames) via the MLLM. This tool allows the agent to use specific prompts to either answer the question directly or acquire complementary cues that are not explicitly exposed by other tools.

\paragraph{Other tools.}
To facilitate downstream reasoning over the extracted cues and relationships, we implement multiple lightweight utility function tools that the agent can call to compute distances, relative directions, heights, and obstructions, etc, with details available in Appendix~\ref{Details of the agent}.

\subsection{Multi-Role Agent Framework: ViSRA}
\label{sec: agent}

\begin{wrapfigure}{r}{0.55\textwidth}
\vspace{-1.2em}
\centering
\begin{minipage}{0.53\textwidth}
\small

\hrule
\vspace{0.35em}
\captionof{algorithm}{\small Multi-role agent inference with spatial tools}
\label{alg:multi_agent}
\vspace{0.35em}
\hrule
\vspace{0.45em}

\begin{algorithmic}[1]
\State \textbf{Input:} Question $q$; video $v$; tool set $\mathcal{T}$; tool schemas $\Sigma$; call budget $B$
\State \textbf{Output:} Final answer $a$
\State \textbf{Notation:} $t\in\mathcal{T}$ denotes a selected tool; $\mathbf{x}$ denotes tool arguments under $\Sigma$; $c=(t,\mathbf{x})$ denotes a tool call; $r$ denotes the raw tool output; $e$ denotes its natural-language interpretation; and $\mathcal{C}$ denotes the call chain.
\State \textbf{Roles:} \textbf{Planner} reads $\Sigma$ and $q$ to produce a tool-selection plan $\pi$. \textbf{Reflector} verifies $\mathcal{C}$ and decides whether to stop or call. \textbf{Executor} executes $c$ by invoking $t$ on $v$ and interprets raw outputs. \textbf{Summarizer} consolidates $\mathcal{C}$ to produce the final $a$.

\State $\pi \gets \textbf{Planner}(q,\mathcal{T},\Sigma)$
\State $n \gets 0;\ \mathcal{C} \gets \emptyset$
\While{$n < B$}
  \State $\textsc{Stop}\ \mid\ c \gets \textbf{Reflector}(q,\pi,\mathcal{C},\Sigma)$
  \If{$\textsc{Stop}$}
    \State \textbf{break}
  \EndIf
  \State $(r,e) \gets \textbf{Executor}(c,v)$
  \State $\mathcal{C} \gets \mathcal{C} \cup \{(c,r,e)\};\ n \gets n+1$
\EndWhile
\State $a \gets \textbf{Summarizer}(q,\mathcal{C})$
\State \Return $a$
\end{algorithmic}

\vspace{0.5em}
\hrule

\end{minipage}
\vspace{-1.0em}
\end{wrapfigure}

To fully and efficiently leverage the spatial tool suite, we instantiate ViSRA as a multi-role agent runtime, as illustrated in Figure~\ref{fig:overview}. The system decomposes planning, control, execution, and answer synthesis into four specialized roles: the \textbf{Planner}, the \textbf{Reflector}, the \textbf{Executor}, and the \textbf{Summarizer}.

Given a question, the Planner parses the query together with the tool schemas and produces a structured execution plan that specifies the required evidence and tool sequence. ViSRA then performs a bounded iterative procedure alternating between the Reflector and the Executor. The Reflector monitors the execution state, verifies whether the currently accumulated evidence is sufficient, and determines the next valid action under the predefined tool interfaces and call budget. The Executor dispatches the selected tool calls to expert models, collects the resulting outputs, and converts them into explicit intermediate observations for subsequent reasoning. This process continues until sufficient evidence has been gathered or the call budget is exhausted, after which the Summarizer integrates the full execution trace to produce the final reasoning and answer grounded in accumulated evidence, rather than relying solely on latent cues (Algorithm~\ref{alg:multi_agent}).

By exposing intermediate states and enforcing an explicit execution protocol, ViSRA reduces complex spatial reasoning to a sequence of verifiable subproblems with modular tool interactions. This design improves controllability, extensibility, and evidence grounding, while preserving a human-aligned reasoning process. An inference example of ViSRA is shown in Figure~\ref{fig:example}.




\begin{table*}[!t]
\centering
\caption{\small \textbf{Evaluation Results on VSI-Bench.}
All numbers are accuracy (\%). We report \textbf{Avg.} (mean accuracy of all numerical and multiple-choice questions). For each base MLLM \textbf{augmented with} ViSRA, we report the absolute gain (pp) for each metric in parentheses after the corresponding score.}

\setlength\tabcolsep{3pt}
\resizebox{\textwidth}{!}{
\begin{tabular}{l|cccc|cccc|c}
\toprule
\multirow{2}{*}{\textbf{Methods}}
& \multicolumn{4}{c|}{\textbf{Numerical Answer}}
& \multicolumn{4}{c|}{\textbf{Multiple-Choice Answer}}
& \multirow{2}{*}{\textbf{Avg.}} \\
\cmidrule(lr){2-5}\cmidrule(lr){6-9}
& Obj. Count & Abs. Dist. & Obj. Size & Room Size
& Rel. Dist. & Rel. Dir. & Route Plan & Appr. Order
& \\
\midrule

\multicolumn{1}{l|}{\textit{Proprietary Models}} 
& & & & & & & & & \\
GPT-4o~\cite{hurst2024gpt}
& $46.2$ & $5.3$ & $43.8$ & $38.2$ & $37.0$ & $41.3$ & $31.5$ & $28.5$ & $34.0$ \\
Gemini-1.5 Flash~\cite{team2024gemini}
& $50.8$ & $33.6$ & $56.5$ & $45.2$ & $48.0$ & $39.8$ & $32.7$ & $59.2$ & $45.7$ \\
Gemini-1.5 Pro~\cite{team2024gemini}
& $49.6$ & $28.8$ & $58.6$ & $49.4$ & $46.0$ & $48.1$ & $42.0$ & $68.0$ & $48.8$ \\
Gemini-2.0 Flash~\cite{team2023gemini}
& $52.4$ & $30.6$ & $66.7$ & $31.8$ & $56.0$ & $46.3$ & $24.5$ & $55.1$ & $45.4$ \\

\midrule
\multicolumn{1}{l|}{\textit{Open-source Base Models}}
& & & & & & & & & \\
Qwen2.5-VL-3B~\cite{bai2025qwen2}
& $24.3$ & $24.7$ & $31.7$ & $22.6$ & $38.3$ & $41.6$ & $26.3$ & $21.2$ & $30.5$ \\
Qwen2.5-VL-7B~\cite{bai2025qwen2}
& $42.9$ & $16.6$ & $52.7$ & $42.2$ & $35.5$ & $39.6$ & $31.4$ & $36.7$ & $37.6$ \\
InternVL3-2B~\cite{chen2024internvl}
& $21.8$ & $24.9$ & $22.0$ & $35.0$ & $33.8$ & $44.2$ & $30.5$ & $7.1$  & $27.5$ \\
InternVL3-8B~\cite{chen2024internvl}
& $23.1$ & $28.7$ & $48.2$ & $39.8$ & $36.7$ & $30.7$ & $29.9$ & $39.6$ & $35.2$ \\
LLaVA-OneVision-7B~\cite{li2024llava}
& $47.7$ & $14.0$ & $47.4$ & $12.3$ & $43.5$ & $42.4$ & $29.4$ & $24.4$ & $35.1$ \\

\midrule
\multicolumn{1}{l|}{\textit{Open-source Post-trained Models}}
& & & & & & & & & \\
Spatial-MLLM-4B~\cite{wu2025spatial}
& $65.3$ & $34.6$ & $63.8$ & $44.2$ & $40.7$ & $44.3$ & $34.5$ & $45.5$ & $47.9$ \\
InternVL3.5-8B~\cite{wang2025internvl3_5}
& $53.7$ & $30.5$ & $60.5$ & $46.3$ & $49.9$ & $42.9$ & $33.0$ & $59.1$ & $48.1$ \\
Qwen3-VL-8B~\cite{qwen3technicalreport}
& $69.8$ & $52.0$ & $76.4$ & $57.0$ & $61.4$ & $52.7$ & $39.2$ & $72.2$ & $62.1$ \\
VLM-3R-7B~\cite{fan2025vlm}
& $70.2$ & $49.4$ & $69.2$ & $67.1$ & $65.4$ & $80.5$ & $45.4$ & $40.1$ & $60.9$ \\
GS-Reasoner~\cite{chen2025reasoning}
& $69.1$ & $61.9$ & $70.0$ & $65.7$ & $65.4$ & $88.9$ & $44.3$ & $52.3$ & $64.7$ \\

\midrule
\multicolumn{1}{l|}{\textit{Ours}}
& & & & & & & & & \\

Qwen2.5-VL-3B + ViSRA
& $39.9$ {\scriptsize \pos{($+15.6$)}}
& $27.3$ {\scriptsize \pos{($+2.6$)}}
& $36.9$ {\scriptsize \pos{($+5.2$)}}
& $30.4$ {\scriptsize \pos{($+7.8$)}}
& $41.1$ {\scriptsize \pos{($+2.8$)}}
& $72.0$ {\scriptsize \pos{($+30.4$)}}
& $40.2$ {\scriptsize \pos{($+13.9$)}}
& $40.6$ {\scriptsize \pos{($+19.4$)}}
& $43.1$ {\scriptsize \pos{($+12.6$)}} \\

Qwen2.5-VL-7B + ViSRA
& $46.4$ {\scriptsize \pos{($+3.5$)}}
& $27.8$ {\scriptsize \pos{($+11.2$)}}
& $60.1$ {\scriptsize \pos{($+7.4$)}}
& $47.7$ {\scriptsize \pos{($+5.5$)}}
& $51.3$ {\scriptsize \pos{($+15.8$)}}
& $71.7$ {\scriptsize \pos{($+32.1$)}}
& $43.8$ {\scriptsize \pos{($+12.4$)}}
& $53.7$ {\scriptsize \pos{($+17.0$)}}
& $52.2$ {\scriptsize \pos{($+14.6$)}} \\

InternVL3-2B + ViSRA
& $38.8$ {\scriptsize \pos{($+17.0$)}}
& $27.8$ {\scriptsize \pos{($+2.9$)}}
& $27.8$ {\scriptsize \pos{($+5.8$)}}
& $34.2$ {\scriptsize \negdiff{($-0.8$)}}
& $49.3$ {\scriptsize \pos{($+15.5$)}}
& $80.4$ {\scriptsize \pos{($+36.2$)}}
& $36.6$ {\scriptsize \pos{($+6.1$)}}
& $17.6$ {\scriptsize \pos{($+10.5$)}}
& $41.4$ {\scriptsize \pos{($+13.9$)}} \\

InternVL3-8B + ViSRA
& $36.6$ {\scriptsize \pos{($+13.5$)}}
& $34.6$ {\scriptsize \pos{($+5.9$)}}
& $58.8$ {\scriptsize \pos{($+10.6$)}}
& $52.7$ {\scriptsize \pos{($+12.9$)}}
& $47.3$ {\scriptsize \pos{($+10.6$)}}
& $62.4$ {\scriptsize \pos{($+31.7$)}}
& $46.9$ {\scriptsize \pos{($+17.0$)}}
& $59.7$ {\scriptsize \pos{($+20.1$)}}
& $50.8$ {\scriptsize \pos{($+15.6$)}} \\

LLaVA-OneVision-7B + ViSRA
& $44.5$ {\scriptsize \negdiff{($-3.2$)}}
& $25.8$ {\scriptsize \pos{($+11.8$)}}
& $51.3$ {\scriptsize \pos{($+3.9$)}}
& $21.9$ {\scriptsize \pos{($+9.6$)}}
& $46.9$ {\scriptsize \pos{($+3.4$)}}
& $66.5$ {\scriptsize \pos{($+24.1$)}}
& $36.1$ {\scriptsize \pos{($+6.7$)}}
& $39.3$ {\scriptsize \pos{($+14.9$)}}
& $45.0$ {\scriptsize \pos{($+9.9$)}} \\

\bottomrule
\end{tabular}
}
\vspace{-1em}
\label{tab:vsibench_results}
\end{table*}

\section{Experiments}
Here, we perform systematic comparisons between ViSRA and baselines supported by experimental evidences. The code will be made available upon acceptance.


\subsection{Experimental Setup}
In our implementation, we deliberately leverage the MLLM’s native perceptual and linguistic capabilities, that is, we let the MLLM select candidate key frames for 2D object detection, answers video/image queries, and makes decisions in four roles.

For 2D object detection, we use Rex-Omni~\cite{jiang2025detect} as the detector and SAM 2~\cite{ravi2024sam} as the tracker.
In 2D object detection (including its use in scene modeling), we uniformly sample $64$ frames from each video. We reduce to $32$ frames to accelerate inference for direct video querying, and  sample at $2$ frames per second (FPS) for cross-frame object tracking. To mitigate drift and failures from overly long tracking chains, we cap each tracking run at $50$ frames and terminate the track if the object is absent in two consecutive frames.
In 3D object detection, we adopt VGGT~\cite{wang2025vggt} as the state-of-the-art foundational geometric model. We use its predicted camera parameters and depth maps to generate point maps for more precise scene modeling. Following the 2D setting, we uniformly sample $64$ frames as input. After constructing the full scene representation, we extract point clouds within the bounding box (bbox) of each selected frame for the efficient 3D representation of the object view.

In our agent framework, all roles are instantiated from the same MLLM and deployed via vLLM~\cite{kwon2023efficient}. We build a self-constructed agent architecture to enable flexible customization. To document tools, we parse each tool function and specify its function schema as a tuple of the function name, input arguments, output format, and function description. The agent prompts and detailed tool descriptions are provided in Appendix~\ref{Details of the agent}.

\subsection{Evaluation on VSI-Bench}
VSI-Bench~\cite{yang2025thinking} comprises over $5,000$ spatially grounded QA pairs collected from egocentric videos in ScanNet~\cite{dai2017scannet}, ScanNet++~\cite{yeshwanth2023scannet++}, and ARKitScenes~\cite{baruch2021arkitscenes}. It covers eight question types, ranging from configurational reasoning and measurement estimation to spatio-temporal reasoning. From the perspective of the answer format, questions fall into two categories: Numerical Answers (NA) and Multiple-Choice Answers (MCA), which are evaluated by Mean Relative Accuracy (MRA) and Accuracy (ACC), respectively. We evaluate the spatial reasoning performance of ViSRA and its baselines on VSI-Bench, following the official evaluation protocol for metric computation.

To assess the efficiency of our approach, we consider $5$ lightweight open-source base models that have not been post-trained on spatial-specific tasks, including Qwen2.5-VL-3B/7B~\cite{bai2025qwen2}, InternVL3-2B/8B~\cite{chen2024internvl}, and LLaVA-OneVision-7B~\cite{li2024llava}. These lightweight base models are more likely to be deployed on edge devices, and thus more exposed to real-world challenges. In addition, we also select $5$ open-source state-of-the-art models (i.e., Spatial-MLLM-4B~\cite{wu2025spatial}, InternVL3.5-8B~\cite{wang2025internvl3_5}, Qwen3-VL-8B~\cite{qwen3technicalreport}, VLM-3R-7B~\cite{fan2025vlm}, and GS-Reasoner~\cite{chen2025reasoning}) which were post-trained on spatial reasoning benchmarks.
As shown in Table~\ref{tab:vsibench_results}, our approach consistently improves the base models, yielding an average gain of over $13\%$ on the final metric. In particular, Qwen2.5-VL-7B reaches $52.2\%$, which is competitive with several proprietary and post-trained models. Across question types, our approach yields the largest gains on categories that strongly depend on genuine spatial understanding, such as relative direction, route planning, and appearance order, highlighting its effectiveness in improving spatial reasoning capabilities.


\begin{wraptable}{r}{0.4\textwidth}
\vspace{-0.8em}
\centering
\caption{\small\textbf{Results on VSI-Bench-Extra.} For each base MLLM augmented with ViSRA, the absolute gain (pp) is shown in parentheses after the corresponding score.}
\label{tab:extra}
\setlength{\tabcolsep}{2pt}
\renewcommand{\arraystretch}{0.95}
\scriptsize
\resizebox{0.4\textwidth}{!}{%
\begin{tabular}{@{}l|ccc|c@{}}
\toprule
\multirow{2}{*}{\textbf{Methods}}
& \textbf{RDB} & \textbf{OO} & \textbf{RDF}
& \multirow{2}{*}{\textbf{Avg.}} \\
& \textbf{(Back)} & \textbf{(Obs.)} & \textbf{(Far)} & \\
\midrule

\multicolumn{5}{l}{\textit{Base Models}} \\
Qwen2.5-VL-3B & 29.3 & 43.8 & 31.3 & 33.5 \\
Qwen2.5-VL-7B & 30.1 & 44.0 & 27.8 & 32.9 \\
InternVL3-8B   & 32.5 & 47.0 & 33.3 & 36.4 \\
\midrule

\multicolumn{5}{l}{\textit{Post-trained Models}} \\
Spatial-MLLM-4B & 30.0 & 35.7 & 14.2 & 27.0 \\
Qwen3-VL-8B     & 32.3 & 45.0 & 40.1 & 37.7 \\
InternVL3.5-8B  & 28.8 & 38.5 & 34.5 & 32.8 \\
\midrule

\multicolumn{5}{l}{\textit{Ours}} \\
Qwen2.5-VL-3B + ViSRA
& $71.5$ {\tiny \pos{($+42.2$)}}
& $48.6$ {\tiny \pos{($+4.8$)}}
& $47.3$ {\tiny \pos{($+16.0$)}}
& $59.0$ {\tiny \pos{($+25.5$)}} \\

Qwen2.5-VL-7B + ViSRA
& $72.0$ {\tiny \pos{($+41.9$)}}
& $56.0$ {\tiny \pos{($+12.0$)}}
& $49.7$ {\tiny \pos{($+21.9$)}}
& $61.8$ {\tiny \pos{($+28.9$)}} \\

InternVL3-8B + ViSRA
& $62.4$ {\tiny \pos{($+29.9$)}}
& $49.5$ {\tiny \pos{($+2.5$)}}
& $47.3$ {\tiny \pos{($+14.0$)}}
& $55.0$ {\tiny \pos{($+18.6$)}} \\

\bottomrule
\end{tabular}%
}
\vspace{-1em}
\end{wraptable}

\begin{table*}[!t]
\centering
\caption{\small \textbf{Additional Results on Other Benchmarks.}
All numbers are accuracy (\%). ViewSpatial denotes camera-relative direction, OST stands for OST-Bench, and MMSI stands for MMSI-Video-Bench. For each base MLLM \textbf{augmented with} ViSRA, the absolute gain (pp) is shown in parentheses after the corresponding score.}
\setlength\tabcolsep{3pt}
\resizebox{\textwidth}{!}{
\begin{tabular}{l|c|cccccc|ccccc}
\toprule
\multirow{2}{*}{\textbf{Methods}}
& \textbf{ViewSpatial}
& \multicolumn{6}{c|}{\textbf{OST}}
& \multicolumn{5}{c}{\textbf{MMSI}} \\
\cmidrule(lr){2-2}\cmidrule(lr){3-8}\cmidrule(lr){9-13}
& Cam.-Rel.-Dir.
& Dir.-Tempo & Dis.-Tempo & Dir.-Est. & Dir.-Judge & Dist.-Judge & Avg.
& Inst.-Scen. & Scen.-Scen. & Cam.-Inst. & Cam.-Scen. & Avg. \\
\midrule

\multicolumn{13}{l}{\textit{Base Models}} \\
Qwen2.5-VL-7B
& $28.8$ & $28.3$ & $44.4$ & $12.8$ & $40.5$ & $45.0$ & $38.5$
& $23.4$ & $26.1$ & $16.6$ & $28.8$ & $23.7$ \\
LLaVA-OneVision-7B
& $26.9$ & $19.2$ & $24.4$ & $23.5$ & $38.1$ & $12.8$ & $25.3$
& $22.1$ & $33.3$ & $21.8$ & $37.6$ & $28.6$ \\
InternVL3-8B
& $28.5$ & $25.5$ & $38.5$ & $13.1$ & $38.2$ & $46.5$ & $37.2$
& $26.0$ & $27.5$ & $20.5$ & $26.3$ & $25.0$ \\

\midrule
\multicolumn{13}{l}{\textit{Ours}} \\
Qwen2.5-VL-7B + ViSRA
& $43.2$ {\scriptsize \pos{($+14.4$)}}
& $66.6$ {\scriptsize \pos{($+38.3$)}}
& $62.3$ {\scriptsize \pos{($+17.9$)}}
& $63.1$ {\scriptsize \pos{($+50.3$)}}
& $70.6$ {\scriptsize \pos{($+30.1$)}}
& $66.3$ {\scriptsize \pos{($+21.3$)}}
& $67.1$ {\scriptsize \pos{($+28.6$)}}
& $27.3$ {\scriptsize \pos{($+3.9$)}}
& $27.5$ {\scriptsize \pos{($+1.4$)}}
& $42.3$ {\scriptsize \pos{($+25.7$)}}
& $37.5$ {\scriptsize \pos{($+8.7$)}}
& $33.9$ {\scriptsize \pos{($+10.2$)}} \\

LLaVA-OneVision-7B + ViSRA
& $46.3$ {\scriptsize \pos{($+19.4$)}}
& $79.9$ {\scriptsize \pos{($+60.7$)}}
& $61.4$ {\scriptsize \pos{($+37.0$)}}
& $72.2$ {\scriptsize \pos{($+48.7$)}}
& $78.1$ {\scriptsize \pos{($+40.0$)}}
& $69.4$ {\scriptsize \pos{($+56.6$)}}
& $73.0$ {\scriptsize \pos{($+47.7$)}}
& $22.1$ {\scriptsize ($+0.0$)}
& $29.0$ {\scriptsize \negdiff{($-4.3$)}}
& $23.1$ {\scriptsize \pos{($+1.3$)}}
& $42.1$ {\scriptsize \pos{($+4.5$)}}
& $29.2$ {\scriptsize \pos{($+0.6$)}} \\

InternVL3-8B + ViSRA
& $45.2$ {\scriptsize \pos{($+16.7$)}}
& $80.8$ {\scriptsize \pos{($+55.3$)}}
& $63.1$ {\scriptsize \pos{($+24.6$)}}
& $71.4$ {\scriptsize \pos{($+58.3$)}}
& $78.5$ {\scriptsize \pos{($+40.3$)}}
& $69.1$ {\scriptsize \pos{($+22.6$)}}
& $73.3$ {\scriptsize \pos{($+36.1$)}}
& $29.9$ {\scriptsize \pos{($+3.9$)}}
& $24.6$ {\scriptsize \negdiff{($-2.9$)}}
& $47.4$ {\scriptsize \pos{($+26.9$)}}
& $35.0$ {\scriptsize \pos{($+8.7$)}}
& $34.5$ {\scriptsize \pos{($+9.5$)}} \\

\bottomrule
\end{tabular}
}
\vspace{-1em}
\label{tab:other_benchmarks}
\end{table*}

\subsection{Evaluation on VSI-Bench-Extra}
\label{sec: extra}

As mentioned in Section~\ref{sec: ob}, we construct VSI-Bench-Extra to preliminarily evaluate the generalization of post-trained MLLMs. VSI-Bench-Extra includes three new question types—relative direction backward, object obstruction, and farthest relative distance, which are either strongly spatial in nature or simply derived from VSI-Bench questions. It contains about 1,600 questions and shares the same video sources as VSI-Bench, with construction details provided in Appendix~\ref{VSI-Bench-Extra}. We evaluate three baseline models, three post-trained models, and our approach applied to the three baselines. As shown in Table~\ref{tab:extra}, post-trained models perform poorly on these unseen questions given videos from VSI-Bench, despite their superior performance on the established VSI-Bench tasks (Table~\ref{tab:vsibench_results}). In contrast, our approach achieves consistently stronger results on VSI-Bench-Extra, with improvements of up to $28.9\%$. In our simple setting of VSI-Bench videos and new questions, experimental results sufficiently prove that ViSRA generalizes better to OOD spatial question types than post-training methods.

\subsection{Evaluation on Other Benchmarks}
To better demonstrate the generalization ability of our method, we expanded our evaluation to a broader set of spatial reasoning benchmarks. Specifically, these benchmarks include ViewSpatial-Bench~\cite{li2025viewspatial}, a benchmark for cross-viewpoint spatial reasoning from human-centered perspectives; OST-Bench~\cite{lin2025ost}, a benchmark for evaluating online spatio-temporal understanding during agent-centric scene exploration; and MMSI-Video-Bench~\cite{lin2025mmsi}, a comprehensive, fully human-annotated benchmark for video-based spatial intelligence in MLLMs. We slightly adjusted the benchmark settings and tool definitions to adapt our method to these benchmarks. For example, for the multi-image setting of ViewSpatial-Bench, we changed the tool input from video to multiple images, and for OST-Bench, we converted the original multi-turn evaluation setting into multiple independent one-turn evaluations. Additionally, in Table~\ref{tab:other_benchmarks}, we only report the task categories that can genuinely benefit from our tools and agent framework.

As shown in Table~\ref{tab:other_benchmarks}, our method significantly improves the performance of base models on these benchmarks. For instance, Qwen2.5-VL-7B achieves performance gains of $14.4\%$, $28.6\%$, and $10.2\%$ on ViewSpatial-Bench, OST-Bench, and MMSI-Video-Bench, respectively. Although the current gains are concentrated in specific task categories, these consistent improvements highlight ViSRA's strong ability to generalize to entirely new data distributions.

\subsection{Ablation Study}



\begin{wraptable}{r}{0.39\textwidth}
\vspace{-1em}
\centering
\small

\captionof{table}{\small \textbf{Ablations of spatial tools on VSI-Bench.}
We vary (i) the number of sampled frames for 2D/3D detection, (ii) the detection model, and (iii) the tracking model.
Metrics are \textbf{MC} (Multiple-Choice), \textbf{Num} (Numerical), and \textbf{Avg.} (their average) in accuracy (\%).
The best setting is frames as \textbf{64}, detector as \textbf{Rex-Omni}, and tracker as \textbf{SAM 2}; its result is shared across ablation groups.}
\label{tab:ablation1}

\setlength\tabcolsep{3pt}
\resizebox{0.36\textwidth}{!}{%
\begin{tabular}{l c c c c}
\toprule
\textbf{Setting} & \textbf{Num./Model} & \textbf{MC} & \textbf{Num} & \textbf{Avg.} \\
\midrule
\multirow{3}{*}{Frames}
& $64$ & $59.2$ & $45.6$ & $52.2$ \\
& $32$ & $55.7$ & $45.5$ & $50.4$ \\
& $16$ & $48.9$ & $44.7$ & $46.7$ \\
\midrule
\multirow{2}{*}{\makecell[l]{Detection\\Model}}
& Rex-Omni & $59.2$ & $45.6$ & $52.2$ \\
& Grounding-DINO & $58.0$ & $42.0$ & $49.0$ \\
\midrule
\multirow{2}{*}{\makecell[l]{Tracking\\Model}}
& SAM 2 & $59.2$ & $45.6$ & $52.2$ \\
& SAM 3 & $59.1$ & $45.0$ & $51.9$ \\
\bottomrule
\end{tabular}%
}

\vspace{0.4em}

\captionof{table}{\small \textbf{Ablation on inference and tool choices for object counting questions.}
We report MRA (\%) on an object-counting subset of VSI-Bench using Qwen2.5-VL-7B as the core MLLM.
\textbf{Inference} denotes whether we directly answer the question with the MLLM (\emph{Direct QA}) or follow our pipeline (\emph{ViSRA}).
\textbf{Clustering} uses either the baseline algorithm DBSCAN~\cite{ester1996density} or our proposed algorithm Constrained Greedy (CG in Appendix~\ref{Clustering Algorithm}).}
\label{tab:ablation2}

\setlength\tabcolsep{4pt}
\resizebox{0.36\textwidth}{!}{%
\begin{tabular}{l c c c c}
\toprule
\textbf{Inference} & \textbf{Detection} & \textbf{Tracking} & \textbf{Clustering} & \textbf{MRA} \\
\midrule
Direct QA & $\times$ & $\times$ & $\times$ & $36.9$ \\
ViSRA & GT & $\times$ & DBSCAN & $56.3$ \\
ViSRA & GT & SAM 2 & DBSCAN & $60.7$ \\
ViSRA & GT & SAM 2 & CG & $80.6$ \\
ViSRA & Rex-Omni & SAM 2 & CG & $55.0$ \\
\bottomrule
\end{tabular}%
}

\vspace{-3em}
\end{wraptable}

\paragraph{Effectiveness of spatial tools.}
To validate the effectiveness of our spatial tools, we conduct ablation studies on (i) the number of sampled frames and (ii) the specific choices of the detection and tracking models.
Specifically, we sample $16$, $32$, or $64$ frames for 2D object detection (and for 3D detection), and evaluate on VSI-Bench with Qwen2.5-VL-7B as the core MLLM.
As shown in Table~\ref{tab:ablation1}, performance improves with increasing sampled frames: using $64$ frames achieves the best average accuracy of $52.2\%$, while reducing to $16$ frames leads to a $5.5\%$ absolute drop.
This highlights the importance of sufficient frame coverage, since frames serve as the primary source of visual and spatial evidence for the entire pipeline.

We further investigate the difference on detection models and tracking models. We first compare two state-of-the-art detection models (i.e., Rex-Omni and Grounding-DINO), with experimental evidence showing that Rex-Omni performs better than Grounding-DINO on both Numerical and Multiple-Choice tasks (Table~\ref{tab:ablation1}). This indicates that Rex-Omni provides more precise per-frame object detection than Grounding-DINO in our setting. We also compare SAM series tracking models (i.e., SAM 2 and SAM 3), obtaining no essential performance gap between them. While SAM 3 added the Promptable Concept Segmentation attribute and improved the Promptable Visual Segmentation capability, a plausible explanation may be that SAM 2 is reliable enough for short-term object tracking in multi-frame videos. 



\paragraph{Potential gains from stronger tools.}
We have partially demonstrated the benefit of decomposing spatial reasoning into subproblems solved by dedicated visual/spatial tools through comparing detection models in Table~\ref{tab:ablation1}. To further understand the benefit of the decomposing mechanism, we conduct additional experiments on the \emph{object counting} subset of VSI-Bench using ground-truth (GT) detection results.
Specifically, we randomly sample about $80$ object-counting questions, manually annotate the GT bboxes on each sampled frame, and use them as the detection outputs in our agent framework ViSRA.
As reported in Table~\ref{tab:ablation2}, performance consistently improves when we add a competitive tracking model and adopt a stronger clustering strategy, eventually reaching the $80.6\%$ accuracy given GT detection results (the second last row)—an absolute gain of $25.6\%$ over using the Rex-Omni detector (the last row).
These results indicate that introducing a state-of-the-art tracking model and clustering algorithm both strengthen the spatial reasoning capability of MLLMs, and improving the detector can yield the largest potential gains if it can attain the near-oracle performance.
Overall, this underscores the controllable and extensible attribute of ViSRA: our agent framework can naturally inherit continual improvements from state-of-the-art models and algorithms without retraining.

\section{Conclusion}
In this work, we revisited the recent push towards 3D spatial intelligence in MLLMs and argued that benchmark-driven post-training, while effective, can blur the line between genuine 3D understanding and task-specific overfitting. 
Under the training-free paradigm, we introduced \textbf{ViSRA}, an inference-time, \textbf{Vi}deo-based \textbf{S}patial \textbf{R}easoning \textbf{A}gent, that enables transferable 3D spatial reasoning by leveraging continual advances in modular and extensible perception models.
Without post-training computational cost and heavy manual curation of datasets, ViSRA delivers stronger performance over baselines on both established spatial benchmarks and OOD tasks. We hope ViSRA complements ongoing efforts on training-free 3D spatial reasoning by offering inference-time and human-aligned tool agents for MLLMs.

\balance
\small{
\bibliography{spatial}  
}
\appendix
\newpage


\begin{algorithm}[t]
\caption{Constrained Greedy Clustering: Views $\rightarrow$ Instances (Single Category)}
\label{alg:view_to_instance_cluster}
\small
\begin{algorithmic}[1]
\Require distance threshold $\varepsilon$; views $\mathcal{V}=\{v_k\}_{k=1}^{K}$
\Statex \hspace{\algorithmicindent}with $v_k=(f_k,\mathbf{b}_k,\mathbf{c}_k)$, where $f_k\in\mathbb{N}$ is the frame index,
\Statex \hspace{\algorithmicindent}$\mathbf{b}_k\in\mathbb{R}^4$ is the 2D bbox, and $\mathbf{c}_k\in\mathbb{R}^3$ is the 3D center
\Statex \hspace{\algorithmicindent}optional tracking partition $\mathcal{T}$ of $\mathcal{V}$
\Ensure instances $\mathcal{I}$, where each instance is a set of views

\Statex \textbf{Build merged points $\mathcal{P}$ and initial clusters $\mathcal{C}$.}
\If{$\mathcal{T}$ is provided}
    \State $\mathcal{P}\gets \emptyset$; $\mathcal{C}\gets \emptyset$
    \For{each track group $G\in\mathcal{T}$}
        \State Construct $p_G$ such that
        \Statex \hspace{\algorithmicindent}$\mathcal{F}(p_G)=\{f(v)\mid v\in G\}$,
        \Statex \hspace{\algorithmicindent}$\mathcal{B}(p_G)=\{\mathbf{b}(v)\mid v\in G\}$,
        \Statex \hspace{\algorithmicindent}$\mathbf{c}(p_G)=\frac{1}{|G|}\sum_{v\in G}\mathbf{c}(v)$,
        \Statex \hspace{\algorithmicindent}$\mathcal{U}(p_G)=G$
        \State $\mathcal{P}\gets \mathcal{P}\cup\{p_G\}$; $\mathcal{C}\gets \mathcal{C}\cup\{\{p_G\}\}$
    \EndFor
\Else
    \State $\mathcal{P}\gets \{p_k\}_{k=1}^{K}$, where each $p_k$ is built from $v_k$
    \Statex \hspace{\algorithmicindent}$\mathbf{c}(p_k)=\mathbf{c}(v_k)$, $\mathcal{F}(p_k)=\{f_k\}$,
    \Statex \hspace{\algorithmicindent}$\mathcal{B}(p_k)=\{\mathbf{b}_k\}$, $\mathcal{U}(p_k)=\{v_k\}$
    \State $\mathcal{C}\gets \{\{p_k\}\}_{k=1}^{K}$
\EndIf

\Statex \textbf{Enumerate candidate point pairs and sort by distance once.}
\State $\mathcal{E}\gets \{(p_i,p_j)\mid p_i,p_j\in\mathcal{P},\, i<j\}$
\State Sort $\mathcal{E}$ in ascending order of $d(p_i,p_j)=\lVert \mathbf{c}(p_i)-\mathbf{c}(p_j)\rVert_2$

\Statex \textbf{Greedy merging under hard constraints.}
\For{each pair $(p_i,p_j)$ in $\mathcal{E}$}
    \If{$d(p_i,p_j)>\varepsilon$}
        \State \textbf{break}
    \EndIf
    \State Let $C_i$ and $C_j$ be the clusters in $\mathcal{C}$ containing $p_i$ and $p_j$, respectively
    \If{$C_i = C_j$}
        \State \textbf{continue} \Comment{already in the same cluster}
    \EndIf
    \State $\mathcal{F}(C_i)\gets \bigcup_{p\in C_i}\mathcal{F}(p)$ and $\mathcal{F}(C_j)\gets \bigcup_{p\in C_j}\mathcal{F}(p)$
    \If{$\mathcal{F}(C_i)\cap \mathcal{F}(C_j)\neq \emptyset$}
        \State \textbf{continue} \Comment{no shared frames within one instance}
    \EndIf
    \State $C\gets C_i\cup C_j$
    \State $\mathcal{C}\gets (\mathcal{C}\setminus\{C_i,C_j\})\cup\{C\}$
\EndFor

\Statex \textbf{Convert clusters to instances over original views.}
\State $\mathcal{I}\gets \{\bigcup_{p\in C}\mathcal{U}(p)\mid C\in\mathcal{C}\}$
\State \Return $\mathcal{I}$
\end{algorithmic}
\end{algorithm}

\begin{table*}[t]
\centering




\begin{minipage}[t]{0.8\textwidth}
\vspace{0pt}
\centering
\captionof{table}{\small \textbf{Latency and Accuracy per Question for Qwen2.5-VL-7B on a Single A100.}
Tool-call indicates amortized tool execution time; agent denotes framework overhead. Avg. is the VSI-Bench accuracy (\%).}
\label{tab:latency_qwen7b}
\setlength{\tabcolsep}{4pt}
\renewcommand{\arraystretch}{0.95}
\scriptsize
\resizebox{\linewidth}{!}{%
\begin{tabular}{c|ccc|c}
\toprule
\textbf{Setting (Frames)} & \textbf{Tool-call (s)} & \textbf{Agent (s)} & \textbf{Total (s)} & \textbf{Avg. (\%)} \\
\midrule
$64$ & $19.96$ & $20.18$ & $40.14$ & $52.2$ \\
$32$ & $11.06$ & $19.61$ & $30.67$ & $50.4$ \\
$16$ & $6.02$  & $20.68$ & $26.70$ & $46.7$ \\
\bottomrule
\end{tabular}%
}
\end{minipage}

\vspace{-1em}
\end{table*}

\section{Limitations}
\label{limitations}
\paragraph{Limitations.}
Despite our initial exploration of the inference-time and human-aligned agent for enhancing the spatial intelligence of MLLMs, substantial opportunities remain for future work. In our experiments, the tool set is fixed and relatively limited; a natural next step is to systematically study how tool scale, diversity, and capability trade-offs influence performance and generalization. Moreover, compared with direct question answering, agent-driven tool invocation can introduce nontrivial latency, underscoring the need for more efficient agent designs. We will show the inference latency of our method statistically in Section~\ref{sec:latency}.

\paragraph{Negative societal impacts.}
Improved video-based spatial reasoning may raise privacy and safety concerns. The ability to infer object locations and spatial relations from videos could be misused for surveillance or tracking. Moreover, in embodied or safety-critical applications, incorrect spatial reasoning caused by imperfect perception tools may lead to unsafe decisions. Careful validation, privacy safeguards, and human oversight are therefore necessary before real-world deployment.

\section{Existing Assets and Licenses}
We use publicly available benchmarks, models, and tool components only for research evaluation. All existing assets are credited through their original papers or official repositories, including VSI-Bench, ViewSpatial-Bench, OST-Bench, MMSI-Video-Bench, Qwen2.5-VL, InternVL, LLaVA-OneVision, Spatial-MLLM, Qwen3-VL, VLM-3R, GS-Reasoner, Rex-Omni, Grounding-DINO, and SAM series models. We follow the corresponding licenses and terms of use of these assets, and do not redistribute their data, model weights, or code unless permitted by their original licenses.

\section{Additional Method Details}

\subsection{Constrained Greedy Clustering}
\label{Clustering Algorithm}
We detail the Constrained Greedy (CG) clustering algorithm used in our 3D object detection tool. 
A key constraint is that an object category cannot yield two distinct views from the same frame within one instance; therefore, we enforce a hard frame-disjoint constraint during clustering. 
Meanwhile, we assume that views of the same physical object tend to have smaller 3D-center distances than views from different objects. 
Based on both the key constraint and the solid assumption, we adopt a CG clustering procedure (Algorithm~\ref{alg:view_to_instance_cluster}): we enumerate all point pairs, sort them by Euclidean distance between their 3D centers pairs, and greedily merge the two clusters associated with a pair if (i) the distance is below $\varepsilon$, (ii) the two points are not already in the same cluster, and (iii) the two clusters have no overlapping frames; otherwise, the pair is skipped.

We further incorporate an optional tracking prior $\mathcal{T}$ as an initialization of $\mathcal{V}$. 
If $\mathcal{T}$ partitions $\mathcal{V}$ into several track groups, we pre-merge each group into a single point whose 3D center is the mean of member 3D centers, while its frames and 2D bounding boxes (bboxes) are stored as sets; the CG then proceeds over these merged points. 
The above algorithm is illustrated for one object category, and we run it across all categories in a loop.

\subsection{Inference Latency}
\label{sec:latency}
Table~\ref{tab:latency_qwen7b} reports the average per-question inference latency under different frame-sampling configurations. As shown, a moderate increase in total latency, from $26.70$s to $40.14$s, leads to a substantial improvement in accuracy, from $46.7\%$ to $52.2\%$. These results indicate that this acceptable trade-off enables our system to achieve training-free adaptability, provide explicit spatial evidence, and seamlessly benefit from advances in stronger perception tools without requiring retraining.

\subsection{Details of ViSRA}
\label{Details of the agent}
\paragraph{Agent prompts.} For each agent role in ViSRA—planner, reflector, executor, and summarizer—we craft a dedicated prompt to guide the MLLM to perform its designated function. The prompts are listed below:

\begin{tcolorbox}[before upper=\raggedright, breakable]
\textbf{Prompt for Planner}\\
\emph{Role.} You are a Planner agent that answers video-based spatial questions.\\
\emph{Input.} You will receive only: (i) \texttt{video\_path}, (ii) \texttt{question}, and (iii) available tool JSON schemas.\\
\emph{Core requirements.}
\begin{itemize}
    \item Decide which tools are needed based on the schemas; do not decide concrete numeric hyperparameters, and focus only on the tool chain.
    \item Return a JSON object with three fields:
    \texttt{plan} (a list of tool names in execution order),
    \texttt{tool\_chain} (a list of objects of the form \texttt{\{"tool": "tool\_name"\}} with no arguments), and
    \texttt{information} (a list of required information types).
\end{itemize}
\emph{Decision rules (evaluated in order).}
\begin{itemize}
    \item \textbf{Appearance / temporal questions:} use 2D detection and sort by frame index.
    Set \texttt{information=["2d\_views","appearance\_sorting"]} and
    \texttt{plan=["tool\_2d\_object\_detection"]}.
    \item \textbf{Object attributes:} for object size, dimensions, or material, use knowledge retrieval.
    Set \texttt{information=["visual\_description","retrieved\_knowledge"]} and
    \texttt{plan=["tool\_knowledge\_retrieval"]}.
    \item \textbf{Numerical scene estimation (critical priority):} if the question asks for room area or absolute distance with explicit units (e.g., meters or square meters), use the direct video query tool.
    Set \texttt{information=["visual\_estimation\_response"]} and
    \texttt{plan=["tool\_video\_image\_query"]}.
    Override rule: even if the wording says ``Measuring from the closest point'', if a physical unit is requested, use this tool rather than the 3D calculation tool.
    \item \textbf{General visual VQA:} for general description, color, action, or atmosphere, use \texttt{tool\_video\_image\_query}.
    \item \textbf{3D spatial reasoning / comparisons:} if the question involves spatial layout, counting, or relative comparison (and is not captured by the previous rule), include
    \texttt{["2d\_views","3d\_center","3d\_bbox"]}.
    For relative-distance comparisons, use
    \texttt{["tool\_2d\_object\_detection", "tool\_object\_3d\_detection", "tool\_calculate\_distance"]}
    with \texttt{information=["closest\_point\_distance"]}; for relative direction, use
    \texttt{["tool\_2d\_object\_detection", "tool\_object\_3d\_detection", "tool\_calculate\_direction"]}
    with \texttt{information=["relative\_direction"]}; otherwise (counting or general layout), use
    \texttt{["tool\_2d\_object\_detection", "tool\_object\_3d\_detection"]}.
\end{itemize}
\emph{Output.} Return JSON only.\\[6pt]

\textbf{Prompt for Reflector (Step-by-step Controller)}\\
\emph{Role.} You are an execution agent that iterates step-by-step.\\
\emph{Input.} A specific instruction for the current step, only the JSON schema for that tool, a PLAN (high-level reference), a HISTORY list of prior tool calls/outputs, and the current \texttt{video\_path} and \texttt{question}.\\
\emph{Task.}
\begin{itemize}
    \item Analyze the previous tool output (if any) in the \texttt{analysis} field.
    \item Execute the current-step instruction.
    If instructed to call a tool, output \texttt{action="call\_tool"} and set \texttt{tool} to the exact name given.
\end{itemize}
\emph{Strict rules.}
\begin{itemize}
    \item Use only arguments explicitly defined in the provided \texttt{TOOL\_SCHEMA}.
    \item Strictly follow behavioral hints, logic, and constraints described in the tool docstring/description.
    \item Do not invent arguments that are not in the schema, and do not reuse arguments from previous steps if they do not belong to the current tool.
    \item For \texttt{video\_path}, use the exact provided string.
    \item For \texttt{output\_2d} or \texttt{tool\_3d\_output}, do not repeat the full heavy content from history; instead pass an empty object \texttt{\{\}} or empty list \texttt{[]}, which the system will inject automatically.
    \item If instructed to finalize, output \texttt{action="final"}.
\end{itemize}
\emph{Output format (strict).} Return exactly one JSON object with the following keys:
\begin{verbatim}
{
  "analysis": "...",
  "action": "call_tool" | "final",
  "tool": "...",        // if action="call_tool"
  "args": {...},        // if action="call_tool"
  "final_answer": "..." // if action="final"
}
\end{verbatim}
\vspace{4pt}

\textbf{Prompt for Executor (Tool Result Interpreter)}\\
\emph{Role.} You are an evidence interpreter (Executor) for a video spatial reasoning agent.\\
\emph{Input.} You will be given the tool JSON schema, tool input arguments, tool output result, and the original question (context only).\\
\emph{Task.}
\begin{itemize}
    \item Generate exactly \textbf{one} natural-language paragraph that describes \textbf{only} the tool output.
    \item Focus on the video content in the returned result (objects, frames, locations, 3D centers).
    \item Cite concrete evidence using exact keys/values from the output (e.g., \texttt{frames=[...]}, bbox entries, \texttt{3d\_center} values).
    \item Do \textbf{not} answer the original question.
\end{itemize}
\emph{Output format (strict).} Return exactly one JSON object and no extra text:
\begin{verbatim}
{
  "result_description": "..."
}
\end{verbatim}
\vspace{4pt}

\textbf{Prompt for Summarizer (Final Answer Integrator)}\\
\emph{Role.} You are a Summarizer agent.\\
\emph{Task.}
\begin{itemize}
    \item \textbf{Analyze evidence, do not narrate process:} ignore the history of which tools were called; focus strictly on \texttt{tool\_output} and \texttt{result\_description}; extract actual values such as 3D coordinates, calculated distances, direction strings, and object counts; then perform explicit logical comparisons over these values to reach the conclusion.
    \item \textbf{Extract the final answer:} for multiple-choice questions, \texttt{final\_answer} must contain only the option letter; for numerical questions, it must contain only the digit(s). Do not include explanation, punctuation, or extra words in \texttt{final\_answer}.
\end{itemize}
\emph{Output format (strict).} Return exactly one JSON object and no extra text:
\begin{verbatim}
{
  "summary": "...",
  "final_answer": "..."
}
\end{verbatim}
\end{tcolorbox}

\paragraph{Description of the tools.}
Tool descriptions are an important part for constructing tool schema, as they guide the agent to select the appropriate tool at each step. We specify each tool’s input, output, and the information it provides. Detailed descriptions are given below:

\begin{tcolorbox}[
  enhanced,
  breakable,
  before upper=\raggedright,
  title=\textbf{Tool Descriptions},
  colback=white,
  colframe=black!50,
  boxrule=0.6pt,
  arc=2pt,
  left=6pt, right=6pt, top=6pt, bottom=6pt
]

\newcommand{\tooldivider}{\par\medskip\hrule\medskip}

\tcbsubtitle{\textbf{2D Object Detection Tool (\texttt{tool\_2d\_object\_detection})}}

\textbf{IMPORTANT:}
\begin{itemize}\setlength{\itemsep}{1pt}
  \item The returned detections are \textbf{category-level views} collected across frames and are \textbf{not} instance-aware; do not use this tool for counting.
  \item When generating the \texttt{objects} argument, the agent must perform noun-phrase extraction over the question and \textbf{must include the reference object} (anchor object) together with the target objects.
\end{itemize}

\textbf{Use this tool for the following question types:}
\begin{itemize}\setlength{\itemsep}{1pt}
  \item Appr. Order. (first-time appearance order)
  \item Rough 2D position questions (left/right in the image)
  \item As the mandatory prerequisite for \texttt{tool\_object\_3d\_detection}
\end{itemize}

\textbf{Special object-parsing examples:}
\begin{itemize}\setlength{\itemsep}{1pt}
  \item ``Which object is closest to the TV?'' $\rightarrow$ \texttt{objects="chair, sofa, stool, tv"}.
  \item ``Is the tv to the left of the sofa?'' $\rightarrow$ \texttt{objects="tv, sofa"}.
\end{itemize}

\textbf{Inputs:}
\begin{itemize}\setlength{\itemsep}{1pt}
  \item \texttt{video\_path}: \texttt{str}. Local video file path (prefer absolute).
  \item \texttt{objects}: \texttt{str}. Comma-separated object category names. Include both the target objects and the reference object.
\end{itemize}

\textbf{Output:}
\begin{itemize}\setlength{\itemsep}{1pt}
  \item A dictionary whose keys are object categories and whose values contain the collected views:
\end{itemize}

\begin{verbatim}
{
  "chair": { "views": [...] },
  "table": { "views": [...] }
}
\end{verbatim}

\tooldivider

\tcbsubtitle{\textbf{3D Object Detection Tool (\texttt{tool\_object\_3d\_detection})}}

\textbf{IMPORTANT:}
\begin{itemize}\setlength{\itemsep}{1pt}
  \item The output is a \textbf{list of physical instances}; unlike the 2D tool, this tool can be used for counting.
  \item Instances are sorted by visibility (\texttt{member\_count}); index \texttt{\_1} denotes the most prominent instance of a category.
\end{itemize}

\textbf{Usage order:}
\begin{itemize}\setlength{\itemsep}{1pt}
  \item Use after \texttt{tool\_2d\_object\_detection}.
  \item \texttt{using\_tracking=True} is required for object counting, because tracking and clustering are used to prevent over-counting caused by camera motion or view changes.
  \item \texttt{aligned\_scene=True} should be enabled for horizontal/vertical relationships, height, front/back/left/right direction, or absolute layout.
\end{itemize}

\textbf{Use this tool for the following question types:}
\begin{itemize}\setlength{\itemsep}{1pt}
  \item Object counting
  \item Relative distance, using \texttt{3d\_center} for Euclidean comparison
  \item Relative direction, using \texttt{3d\_center} together with \texttt{aligned\_scene=True}
  \item Route planning
\end{itemize}

\textbf{Inputs:}
\begin{itemize}\setlength{\itemsep}{1pt}
  \item \texttt{video\_path}: \texttt{str}. Local video file path (prefer absolute).
  \item \texttt{output\_2d}: the exact dictionary returned by \texttt{tool\_2d\_object\_detection(...)}.
  \item \texttt{using\_tracking}: \texttt{bool}. Default \texttt{False}. Set to \texttt{True} for counting.
  \item \texttt{aligned\_scene}: \texttt{bool}. Default \texttt{False}. Set to \texttt{True} for direction/relation tasks that require a real-world frame.
\end{itemize}

\textbf{Output:}
\begin{itemize}\setlength{\itemsep}{1pt}
  \item A list of instance dictionaries in the following format:
\end{itemize}

\begin{verbatim}
[
  {
    "instance_id": "chair_1",
    "category": "chair",
    "3d_center": [x, y, z],
    "bbox_3d": [x1, y1, z1, x2, y2, z2]
  },
  {
    "instance_id": "chair_2",
    "category": "chair",
    "3d_center": [...],
    "bbox_3d": [...]
  }
]
\end{verbatim}

\tooldivider

\tcbsubtitle{\textbf{Knowledge Retrieval Tool (\texttt{tool\_knowledge\_retrieval})}}

\textbf{Purpose:}
\begin{itemize}\setlength{\itemsep}{1pt}
  \item Retrieve object-/room-level size statistics and descriptions as spatial priors for estimation tasks.
\end{itemize}

\textbf{When to use:}
\begin{itemize}\setlength{\itemsep}{1pt}
  \item For underdetermined estimation questions (e.g., object size estimation, room size estimation) where priors beyond visual evidence are needed.
\end{itemize}

\textbf{Inputs:}
\begin{itemize}\setlength{\itemsep}{1pt}
  \item \texttt{query} (string): question text or distilled caption.
  \item \texttt{top\_k} (int, default: 5): number of entries to return.
\end{itemize}

\textbf{Output:}
\begin{itemize}\setlength{\itemsep}{1pt}
  \item Top-$k$ retrieved entries (e.g., $k=5$), where each entry is a single \texttt{str} containing size statistics and a short description, in the following format:
\end{itemize}

\begin{verbatim}
{
  "query": "...",
  "top_k": 5,
  "entries": [
    "entry string 1",
    "entry string 2",
    ...
  ]
}
\end{verbatim}

\tooldivider

\tcbsubtitle{\textbf{Video/Image Query Tool (\texttt{tool\_video\_image\_query})}}

\textbf{Purpose:}
\begin{itemize}\setlength{\itemsep}{1pt}
  \item Directly query the video (or a selected key frame) via the MLLM, returning a \textbf{natural-language response}.
  \item Acquire complementary cues that are not explicitly exposed by other tools.
\end{itemize}

\textbf{When to use:}
\begin{itemize}\setlength{\itemsep}{1pt}
  \item For general description, visual attributes, action recognition, and estimation-style questions that need direct semantic or metric cues beyond structured 2D/3D outputs.
\end{itemize}

\textbf{Inputs:}
\begin{itemize}\setlength{\itemsep}{1pt}
  \item \texttt{video\_path} (string): local video file path (prefer absolute).
  \item \texttt{prompt} (string): the specific query prompt.
  \item \texttt{query\_type} (string, default: \texttt{"video"}): one of \texttt{\{"video","image"\}}.
  \item \texttt{frame\_idx} (int, default: \texttt{-1}): required when \texttt{query\_type="image"}, specifying the queried frame index.
\end{itemize}

\textbf{Critical prompt templates.}
For room-size and absolute-distance estimation, the tool description in \texttt{tool.py} requires the agent to use the following exact prompt templates (replacing \texttt{\{question\}} with the original question):
\begin{verbatim}
1. Room size / area estimation
prompt = (
    "You are an expert at estimating room size (area) from videos.\n"
    "Use the visual information in the video to answer the user's question.\n"
    "Return ONLY a single best numerical estimate (integer or decimal) in square meters.\n"
    "Output format (STRICT): <answer>NUMBER</answer>\n"
    "- Do NOT output units.\n"
    "- Do NOT output a range.\n"
    "- Do NOT output any explanation.\n\n"
    f"QUESTION: {question}\n"
)

2. Distance estimation (between two objects)
prompt = (
    "You are an expert at estimating REAL-WORLD distance between two objects from videos.\n"
    "Use the visual information in the video to answer the user's question.\n"
    "Return ONLY a single best numerical estimate (integer or decimal) in meters.\n"
    "Output format (STRICT): <answer>NUMBER</answer>\n"
    "- Do NOT output units.\n"
    "- Do NOT output a range.\n"
    "- Do NOT output any explanation.\n\n"
    f"QUESTION: {question}\n"
)
\end{verbatim}
An analogous pair of templates is used in \texttt{tool\_multi\_image\_query}, with ``from videos'' replaced by ``from multiple images''.

\textbf{Output:}
\begin{itemize}\setlength{\itemsep}{1pt}
  \item A dict containing a direct natural-language \texttt{response}:
\end{itemize}

\begin{verbatim}
{
  "video_path": "...",
  "query_type": "video" | "image",
  "frame_idx": <int or null>,
  "prompt": "...",
  "response": "natural language text"
}
\end{verbatim}

\tooldivider

\tcbsubtitle{\textbf{Distance Computation Tool (\texttt{tool\_calculate\_distance})}}

\textbf{Purpose:}
\begin{itemize}\setlength{\itemsep}{1pt}
  \item Calculate Euclidean distance between the 3D centers of object instances.
  \item The resulting distances are relative to the scene scale rather than guaranteed metric distances.
\end{itemize}

\textbf{Inputs:}
\begin{itemize}\setlength{\itemsep}{1pt}
  \item \texttt{tool\_3d\_output}: the list returned by \texttt{tool\_object\_3d\_detection(...)}.
  \item \texttt{video\_path}: absolute path to the video.
  \item \texttt{reference\_instance}: \texttt{str}. It must be a specific instance ID, such as \texttt{tv\_1}, rather than a category name.
  \item \texttt{target\_instances}: \texttt{str}. A comma-separated list of specific instance IDs.
\end{itemize}

\textbf{Output:}
\begin{verbatim}
{
  "video_path": "...",
  "result": {
    "reference_instance": "tv_1",
    "target_instances": ["chair_1", "stool_1", "sofa_1", "stove_1"],
    "distances": {"chair_1": 0.27, ...},
    "unit": "relative"
  }
}
\end{verbatim}

\tooldivider

\tcbsubtitle{\textbf{Relative Direction Computation Tool (\texttt{tool\_calculate\_direction})}}

\textbf{Purpose:}
\begin{itemize}\setlength{\itemsep}{1pt}
  \item Compute the relative direction of a target object under an egocentric reference frame defined by a standing location (\texttt{stand}) and a facing direction (\texttt{face}).
\end{itemize}

\textbf{When to use:}
\begin{itemize}\setlength{\itemsep}{1pt}
  \item For relative-direction questions once 3D centers are available (optionally stabilized by scene modeling).
\end{itemize}

\textbf{Inputs:}
\begin{itemize}\setlength{\itemsep}{1pt}
  \item \texttt{tool\_3d\_output}: the list returned by \texttt{tool\_object\_3d\_detection(...)}.
  \item \texttt{video\_path}: absolute path to the video.
  \item \texttt{stand\_instance}: \texttt{str}. A specific instance ID, e.g., \texttt{stove\_1}. If multiple same-category objects exist, the implementation prefers the most prominent instance (suffix \texttt{\_1}) unless the question implies otherwise.
  \item \texttt{face\_instance}: \texttt{str}. A specific instance ID, e.g., \texttt{sofa\_1}.
  \item \texttt{target\_instance}: \texttt{str}. A specific instance ID, e.g., \texttt{tv\_1}.
\end{itemize}

\textbf{Output:}
\begin{verbatim}
{
  "video_path": "...",
  "result": {
    "stand_instance": "stove_1",
    "face_instance": "sofa_1",
    "target_instance": "tv_1",
    "direction": "front-left|front-right|back-left|back-right",
    "evidence": {...}
  }
}
\end{verbatim}

\tooldivider

\tcbsubtitle{\textbf{Height Comparison Tool (\texttt{tool\_compare\_height})}}

\textbf{Purpose:}
\begin{itemize}\setlength{\itemsep}{1pt}
  \item Compare heights of two objects by comparing their \texttt{z}-axis values.
\end{itemize}

\textbf{IMPORTANT:}
\begin{itemize}\setlength{\itemsep}{1pt}
  \item Requires an \textbf{aligned scene}: 3D outputs must be produced after calling \texttt{tool\_scene\_modeling} so that the \texttt{z}-axis corresponds to real-world vertical direction.
\end{itemize}

\textbf{Inputs:}
\begin{itemize}\setlength{\itemsep}{1pt}
  \item \texttt{tool\_3d\_output}: the list returned by \texttt{tool\_object\_3d\_detection(...)} with alignment applied.
  \item \texttt{video\_path}: must match \texttt{tool\_3d\_output["video\_path"]}.
  \item \texttt{instance\_a}: \texttt{str}. \textbf{MUST} be an \textbf{INSTANCE NAME}.
  \item \texttt{instance\_b}: \texttt{str}. \textbf{MUST} be an \textbf{INSTANCE NAME}.
\end{itemize}

\textbf{Output:}
\begin{verbatim}
{
  "video_path": "...",
  "result": {
    "instance_a": "stool_1",
    "instance_b": "table_1",
    "z_a": 1.23,
    "z_b": 0.87,
    "relation": "a_higher|b_higher|equal"
  }
}
\end{verbatim}

\tooldivider

\tcbsubtitle{\textbf{Object Obstruction Checking Tool (\texttt{tool\_calculate\_obstruction})}}

\textbf{Purpose:}
\begin{itemize}\setlength{\itemsep}{1pt}
  \item Check whether an obstruction object lies on the route (straight-line segment) from a source object to a destination object, using 3D centers.
\end{itemize}

\textbf{Inputs:}
\begin{itemize}\setlength{\itemsep}{1pt}
  \item \texttt{tool\_3d\_output}: the list returned by \texttt{tool\_object\_3d\_detection(...)}.
  \item \texttt{video\_path}: must match \texttt{tool\_3d\_output["video\_path"]}.
  \item \texttt{source\_instance}: \texttt{str}. \textbf{MUST} be an \textbf{INSTANCE NAME}.
  \item \texttt{destination\_instance}: \texttt{str}. \textbf{MUST} be an \textbf{INSTANCE NAME}.
  \item \texttt{obstruction\_instance}: \texttt{str}. \textbf{MUST} be an \textbf{INSTANCE NAME}.
\end{itemize}

\textbf{Output:}
\begin{verbatim}
{
  "video_path": "...",
  "result": {
    "source_instance": "table_1",
    "destination_instance": "sofa_1",
    "obstruction_instance": "stool_1",
    "is_obstruction": true,
    "evidence": {
      "source_center": {"x":..., "y":..., "z":...},
      "destination_center": {"x":..., "y":..., "z":...},
      "obstruction_center": {"x":..., "y":..., "z":...},
      "t": 0.42,
      "closest_point": {"x":..., "y":..., "z":...},
      "distance_to_segment": 0.18,
      "threshold": 0.25
    }
  }
}
\end{verbatim}

\end{tcolorbox}

\section{Details of VSI-Bench-Extra}
\label{VSI-Bench-Extra}
As mentioned in Section~\ref{sec: ob} and~\ref{sec: extra}, we design three new question types to evaluate MLLMs' out-of-distribution (OOD) spatial reasoning capability: \textbf{relative direction backward}, \textbf{object obstruction}, and \textbf{relative distance farthest}. \textbf{Relative direction backward} modifies VSI-Bench relative-direction prompt from `facing \dots'' to `with my back to \dots,'' and correspondingly flips the ground-truth (GT) answer. \textbf{Object obstruction} asks whether (and which) an object obstructs the route from object A to object B. \textbf{Relative distance farthest} asks for the farthest object, in contrast to the nearest-object formulation used in VSI-Bench.

For \textbf{relative direction backward}, we directly reuse the medium and hard subsets of VSI-Bench relative-direction questions, rewrite the questions, and reverse the answers as ground truth, resulting in $751$ questions. For \textbf{object obstruction} and \textbf{relative distance farthest}, we follow the VSI-Bench construction pipeline: we leverage GT bounding boxes from the scan datasets ARKitScenes, ScanNet, and ScanNet++, filter out ambiguous annotations, and generate questions using templates with human verification. This yields $400$ \textbf{object obstruction} questions and $444$ \textbf{relative distance farthest} questions. The templates used to generate these three types are as follows:

\begin{tcolorbox}[before upper=\raggedright]
\textbf{Relative Direction Backward}\\
If I am standing by the \textcolor{red}{\{positioning object\}} and with my back to the \textcolor{red}{\{orienting object\}} (facing directly away from it), is the \textcolor{red}{\{querying object\}} to my front-left, front-right, back-left, or back-right?\\
Directions refer to the quadrants of a Cartesian plane (assuming I am standing at the origin and facing the positive y-axis).\\[4pt]

\textbf{Object Obstruction}\\
If I am standing by the \textcolor{red}{\{positioning object\}} and facing the \textcolor{red}{\{orienting object\}}, which object is there as an obstruction when I walk straight to the \textcolor{red}{\{orienting object\}}?\\[4pt]

\textbf{Relative Distance Farthest}\\
Measuring from the closest point of each object, which of these objects (\textcolor{red}{\{choice a\}}, \textcolor{red}{\{choice b\}}, \textcolor{red}{\{choice c\}}, \textcolor{red}{\{choice d\}}) is the farthest from the \textcolor{red}{\{category\}}?
\end{tcolorbox}


\section{Qualitative Experimental Results}
\label{app:qualitative_results}
\begin{figure*}[t]
  \begin{center}
    \includegraphics[width=\textwidth]{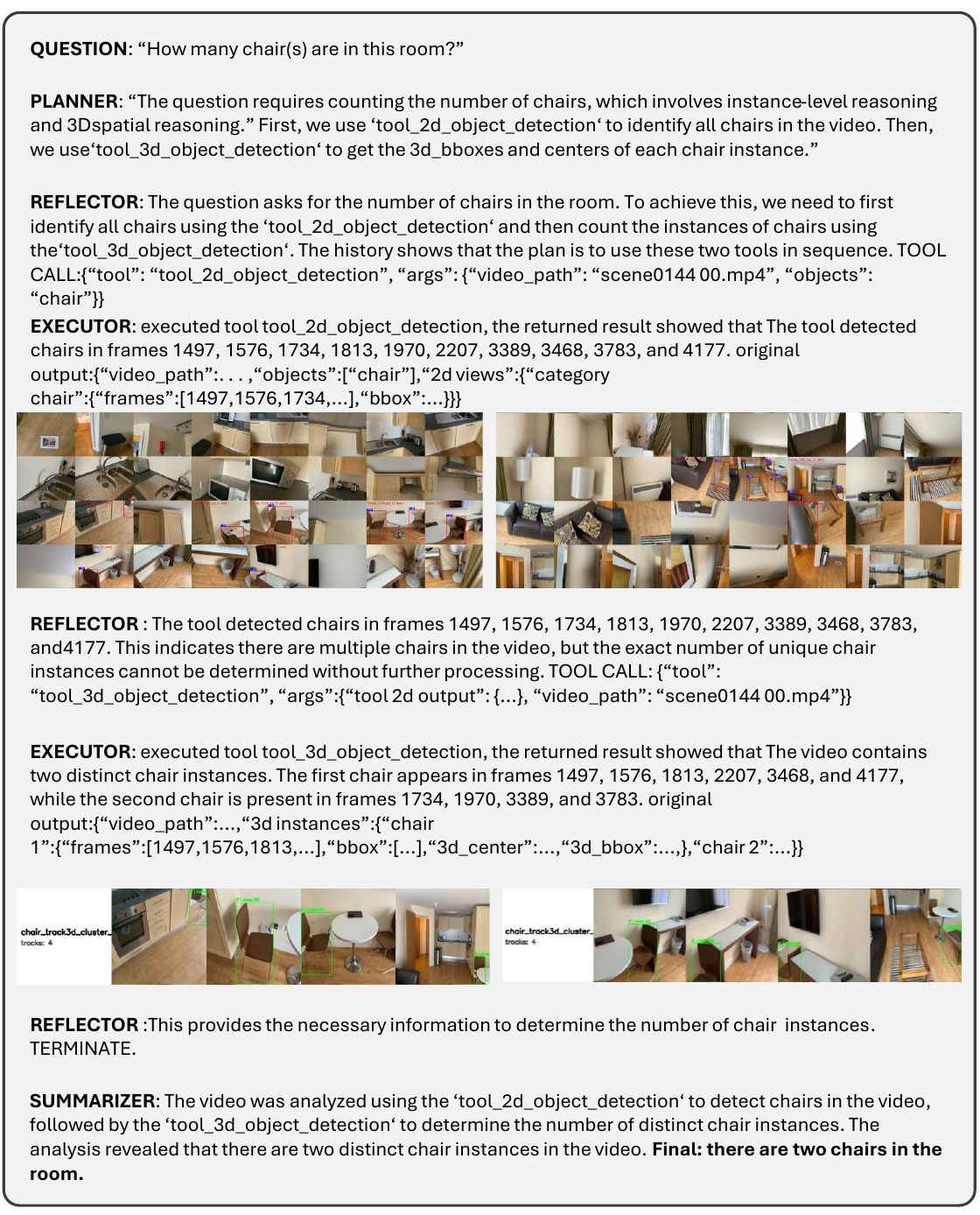}
    \caption{An example of ViSRA solving an object-counting question correctly.}
    \label{fig:appendix qualitative count}
  \end{center}
\end{figure*}

\begin{figure*}[t]
  \begin{center}
    \includegraphics[width=\textwidth]{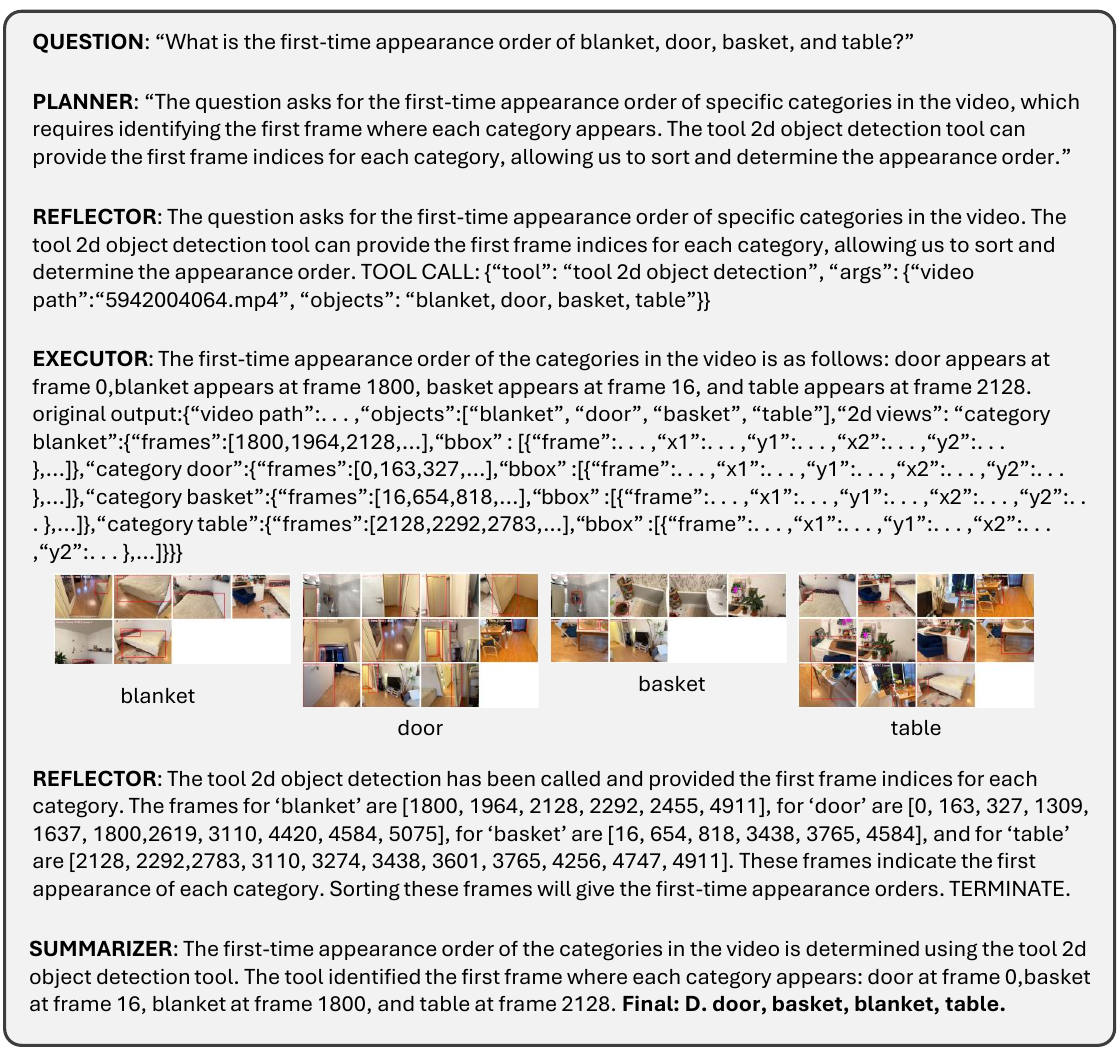}
    \caption{An example of ViSRA solving an appearance-order question correctly.}
    \label{fig:appendix qualitative order}
  \end{center}
\end{figure*}

\begin{figure*}[t]
  \begin{center}
    \includegraphics[width=\textwidth]{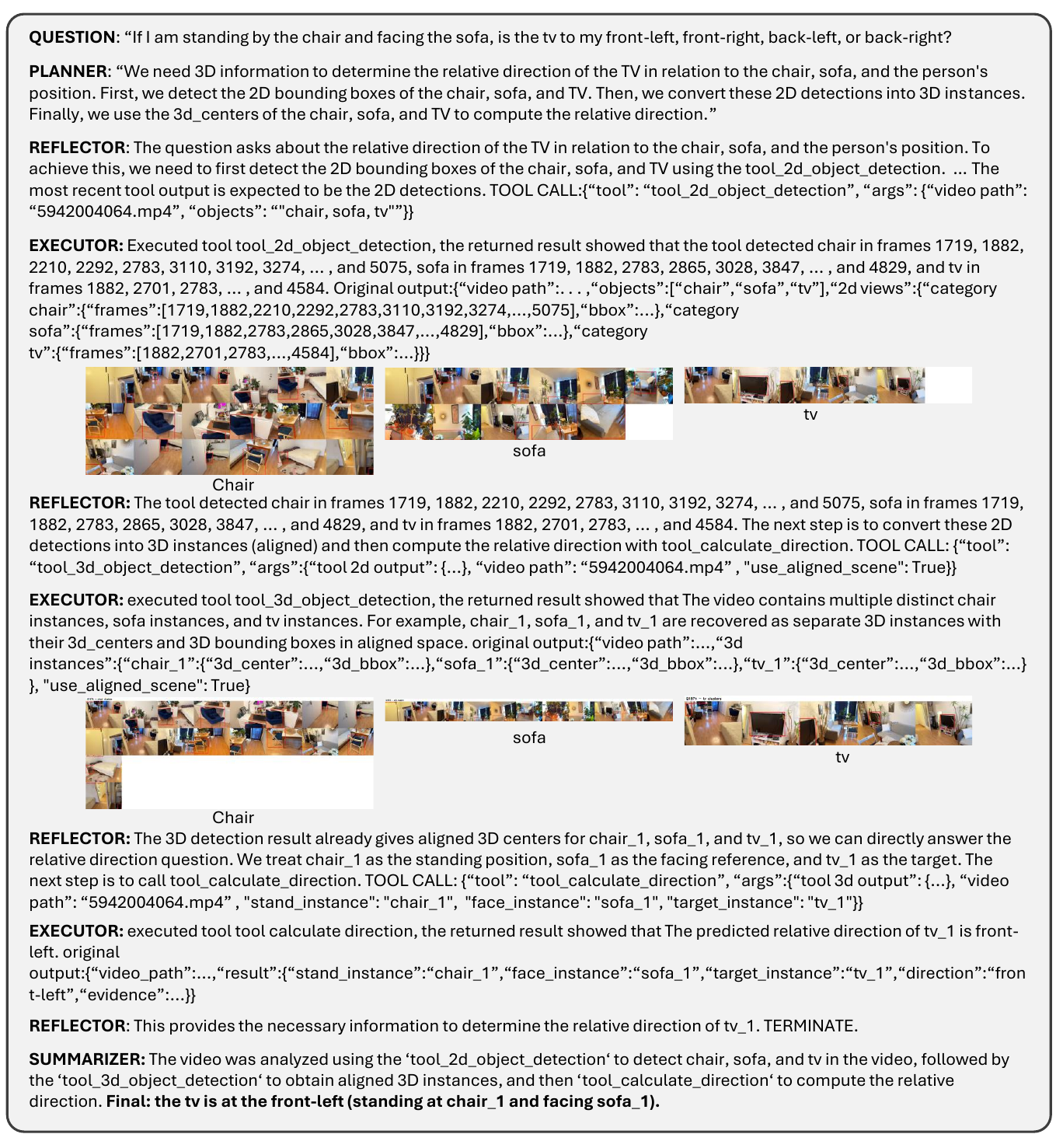}
    \caption{An example of ViSRA solving a relative-direction question correctly.}
    \label{fig:appendix qualitative dir}
  \end{center}
\end{figure*}

\begin{figure*}[t]
  \begin{center}
    \includegraphics[width=\textwidth]{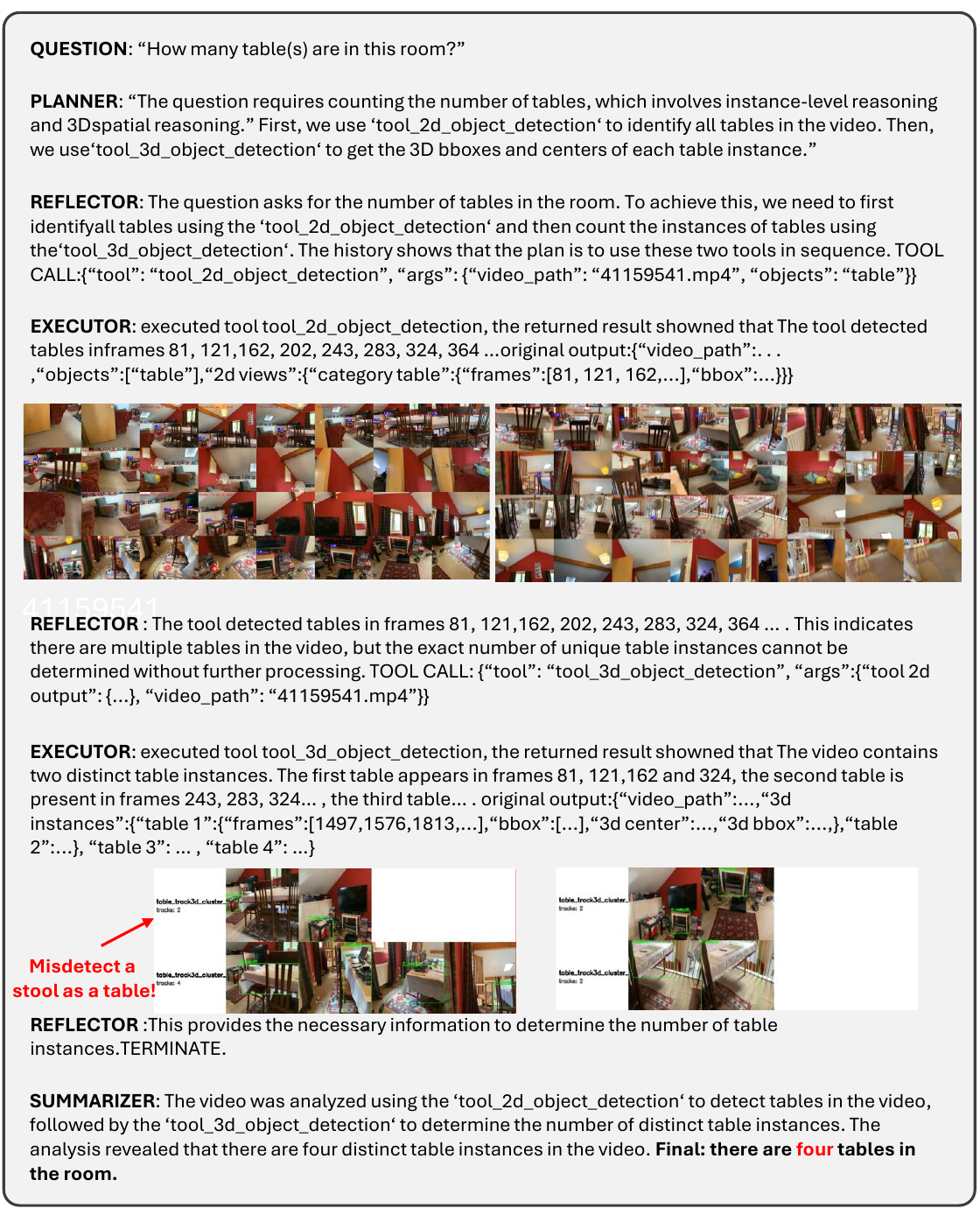}
    \caption{An example of ViSRA solving an object-counting question incorrectly due to misdetecting a stool as a table.}
    \label{fig:appendix qualitative count wrong}
  \end{center}
\end{figure*}

Here we present additional qualitative results, including both successful and failed cases, together with their explicit intermediate spatial representations shown from Figure~\ref{fig:appendix qualitative count} to~\ref{fig:appendix qualitative count wrong}. ViSRA is able to produce correct answers while generating human-interpretable intermediate reasoning outputs by invoking spatial tools. However, it may also fail due to limitations of the underlying expert models (e.g., detectors), as illustrated in Figure~\ref{fig:appendix qualitative count wrong}.

\section{Generation of Ground-truth Cognitive Maps}
\label{app:cognitive}
\begin{figure*}[t]
  \begin{center}
    \includegraphics[width=\textwidth]{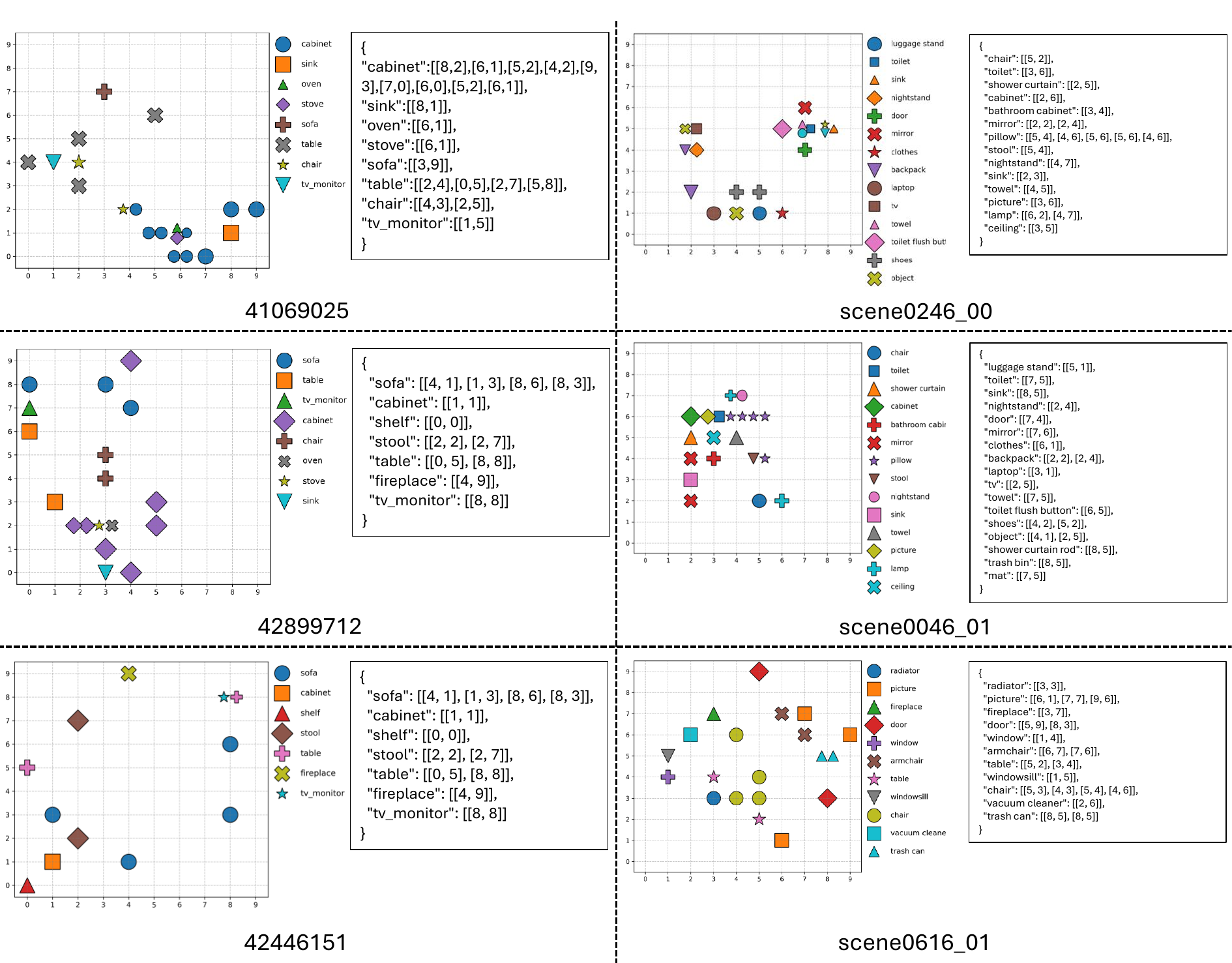}
    \caption{Visualized and textual examples of ground-truth cognitive maps generated from 3D annotations. The three on the left panel are from ARKitScenes, and the three on the right panel are from ScanNet.}
    \label{fig:appendix cogmap}
  \end{center}
\end{figure*}

In Section~\ref{sec: ob}, we constructed experiments to evaluate the ineffectiveness of cognitive maps for spatial intelligence. Specifically, we used GT 3D object centers and bboxes in 3D datasets to generate $81$ cognitive maps for a subset of $779$ questions. We employed a 10×10 grid with normalized scene length while maintaining the x–y aspect ratio for a scene, with the JSON format adhering to the VSI-Bench specification~\cite{yang2025thinking}. In the visualization, objects mapped to the same coordinate were displayed with lateral offsets to prevent overlap (Figure~\ref{fig:appendix cogmap}).

\begin{figure*}[t]
  \begin{center}
    \includegraphics[width=\textwidth]{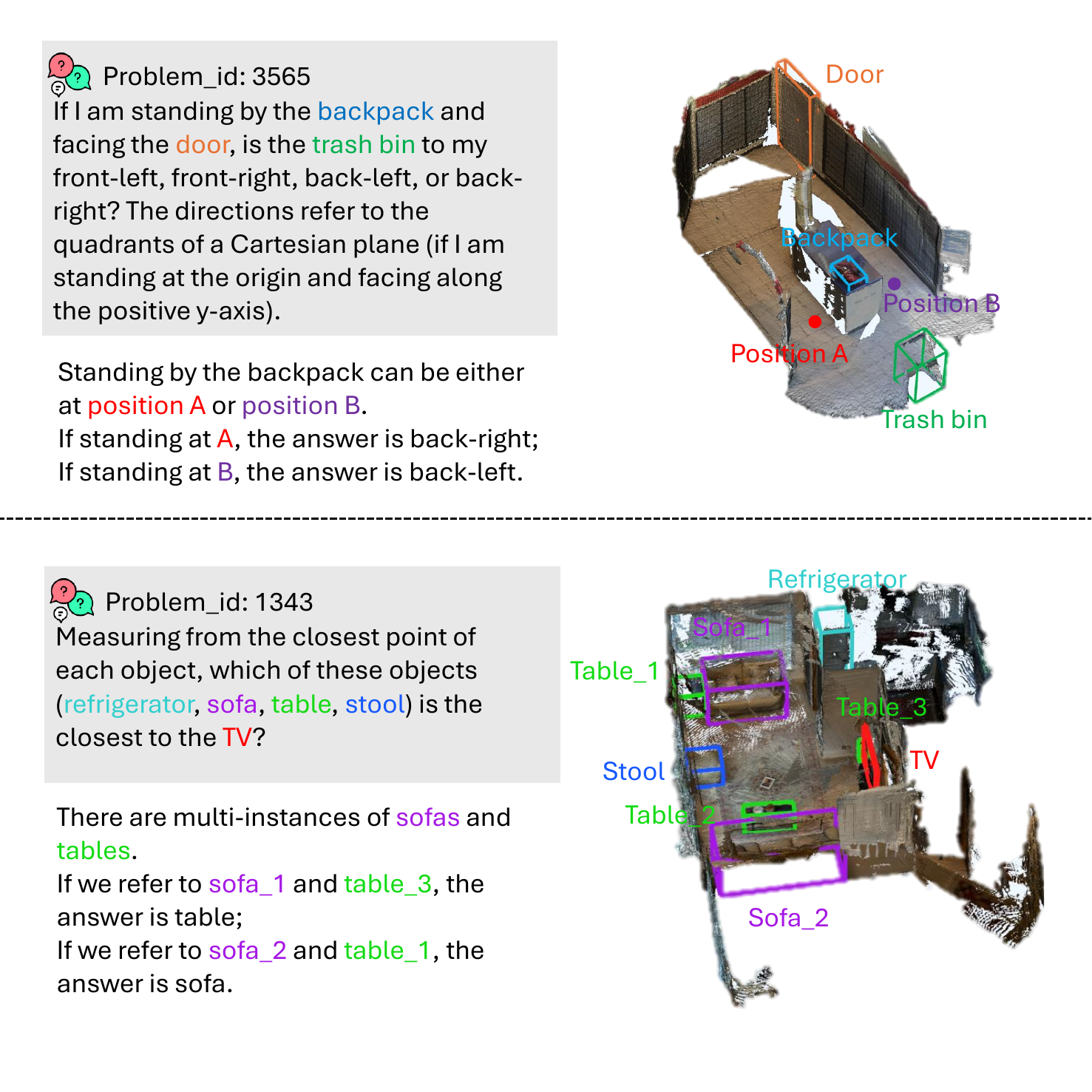}
    \caption{Two examples in VSI-Bench where ambiguous referring expressions lead to ambiguous answers.}
    \label{fig:bad bench}
  \end{center}
\end{figure*}

\section{Limitations of Existing Benchmarks}
Existing benchmarks can contain unsure or wrong answers in their datasets. For example, the problem $3565$ in VSI-Bench asks "If I am standing by the backpack and facing the door, ... ", which is unclear from the video (The upper half of Figure~\ref{fig:bad bench}) as a person can stand by either the left or right of the backpack. Therefore, both B and C appear plausible, but the GT answer is C. 
Another example is that the problem $1343$ involves multi-instance categories (e.g., multiple sofas and tables), leading to different answers depending on which instance is selected. (The lower half of Figure~\ref{fig:bad bench}).
In conclusion, we believe that most established spatial reasoning benchmarks can have similar quality issues unless each question-answer pair together with the video are manually inspected very carefully.

\clearpage

\newpage
\newpage
\end{document}